\documentclass[letterpaper]{article} 
\usepackage{aaai2026}  
\usepackage{times}  
\usepackage{helvet}  
\usepackage{courier}  
\usepackage[hyphens]{url}  
\usepackage{graphicx} 
\usepackage{natbib}  
\usepackage{caption} 
\usepackage{algorithm}
\usepackage{algorithmic}
\usepackage{amsmath}
\usepackage{amsthm}
\usepackage{amsfonts}
\theoremstyle{definition}
\newcommand{\cE}{\mathcal{E}}
\newcommand{\xb}{\mathbf{x}}
\newcommand{\vb}{\mathbf{v}}
\newcommand{\ub}{\mathbf{u}}
\newcommand{\ob}{\mathbf{o}}
\newcommand{\Ab}{\mathbf{A}}
\newcommand{\ab}{\mathbf{a}}
\newcommand{\Ib}{\mathbf{I}}
\newcommand{\pzero}{p_{0}(\xb_0)}
\newcommand{\qzero}{q_{0}(\xb_0)}
\newcommand{\ut}{\ub_{t}(\xb)}
\newcommand{\uthat}{\hat{\ub}_{t}(\xb)}

\newcommand{\pt}{p_{t}(\xb)}
\newcommand{\qt}{q_{t}(\xb)}
\newcommand{\ptzero}{p_{t0}(\xb|\xb_0)}
\newcommand{\qtzero}{q_{t0}(\xb|\xb_0)}
\newcommand{\utzero}{\ub_{t0}(\xb|\xb_0)}

\newtheorem{theorem}{Theorem}
\newtheorem{corollary}{Corollary}
\newtheorem{definition}{Definition}
\newtheorem{assumption}{Assumption}

\usepackage{colortbl} 
\usepackage{multirow}
\makeatletter
\newcommand{\thickhline}{%
    \noalign {\ifnum 0=`}\fi \hrule height 1pt
    \futurelet \reserved@a \@xhline
}
\makeatother
\usepackage[export]{adjustbox}
\usepackage{enumitem}
\usepackage{bm}
\usepackage{xcolor}
\usepackage{listings}
\definecolor{mygray}{gray}{.9}
\usepackage{subcaption} 
\usepackage{makecell}
\usepackage{booktabs} 

\usepackage{newfloat}
\usepackage{listings}
\DeclareCaptionStyle{ruled}{labelfont=normalfont,labelsep=colon,strut=off} 
\floatstyle{ruled}
\newfloat{listing}{tb}{lst}{}
\floatname{listing}{Listing}
\nocopyright 

\title{
Balancing Signal and Variance: Adaptive Offline RL Post-Training\\for VLA Flow Models
}
\author{
    Hongyin Zhang\textsuperscript{\rm 1}, 
    Shiyuan Zhang\textsuperscript{\rm 2}, 
    Junxi Jin\textsuperscript{\rm 1}, 
    Qixin Zeng\textsuperscript{\rm 1}, 
    Yifan Qiao\textsuperscript{\rm 3}, 
    Hongchao Lu\textsuperscript{\rm 1}, 
    Donglin Wang\textsuperscript{\rm 1}\thanks{Corresponding author. wangdonglin@westlake.edu.cn}
}
\affiliations{
    \textsuperscript{\rm 1}
    Westlake University, Hangzhou, China \\
    \textsuperscript{\rm 2}
    University of California, Los Angeles, USA \\
    \textsuperscript{\rm 3}
    Xi'an Jiaotong University, Xi'an, China

}

\usepackage{bibentry}

\begin{document}

\maketitle

\begin{abstract}
Vision-Language-Action (VLA) models based on flow matching have shown excellent performance in general-purpose robotic manipulation tasks. 
However, the action accuracy of these models on complex downstream tasks is unsatisfactory. 
One important reason is that these models rely solely on the post-training paradigm of imitation learning, which makes it difficult to have a deeper understanding of the distribution properties of data quality, which is exactly what Reinforcement Learning (RL) excels at. 
In this paper, we theoretically propose an offline RL post-training objective for VLA flow models and induce an efficient and feasible offline RL fine-tuning algorithm $--$ \textbf{A}daptive \textbf{R}einforced \textbf{F}low \textbf{M}atching (\textbf{ARFM}). 
By introducing an adaptively adjusted scaling factor in the VLA flow model loss, we construct a principled bias-variance trade-off objective function to optimally control the impact of RL signal on flow loss. 
ARFM adaptively balances RL advantage preservation and flow loss gradient variance control, resulting in a more stable and efficient fine-tuning process.
Extensive simulation and real-world experimental results show that ARFM exhibits excellent generalization, robustness, few-shot learning, and continuous learning performance.
\end{abstract}


\section{Introduction}
In recent years, the rapid development of vision-language-action (VLA) models has enabled robots to perform a variety of manipulation tasks based on multi-modal perception. 
Large-scale pre-training systems have demonstrated the feasibility of learning general policies from internet-level multi-modal data and have performed well in real-world robotic manipulation tasks, such as RT-1~\cite{rt_1}, RT-2~\cite{rt_2}, PaLM-E~\cite{palm_e}, Octo~\cite{octo}, OpenVLA~\cite{openvla}, $\pi_{0}$~\cite{pi_0}.
In particular, $\pi_0$ is a VLA flow policy model based on trajectory vector fields, which utilizes trajectory-level denoising modeling mechanism to achieve parallel and efficient policy learning and achieves state-of-the-art performance in realistic manipulation tasks.

Although pre-trained VLA models show good generalizability, fine-tuning on downstream tasks based on the imitation learning paradigm still faces great challenges in terms of action accuracy~\cite{rt_2,black2023zero,li2024evaluating}.
In this scenario, VLA models that rely solely on behavior cloning or flow matching may not be able to effectively exploit the quality distribution structure in the training data. 
Some works~\cite{mark2024policy,zhai2024fine,zhao2025more,guo2025improving,reinbot} resort to offline Reinforcement Learning (RL) in the hope of mining deeper and fine-grained data quality features to achieve more efficient VLA model fine-tuning. 
Recently, an algorithm called ReinboT~\cite{reinbot} was proposed, which attempts to introduce RL return-to-go as a fine-grained goal to guide the fine-tuning of the VLA model.
However, we observed that this approach has limited performance in VLA flow models (such as $\pi_0$), see the experimental section for details. 
We infer that this is because VLA flow models the entire action trajectory distribution through a vector field. 
During the inference phase of the VLA flow model, the maximized return-to-go can only control the generation of the trajectory vector field, thereby indirectly and inefficiently guiding the final action prediction. 
Therefore, how to effectively perform offline RL fine-tuning of VLA flow models remains underexplored.

To this end, we propose \textbf{A}daptive \textbf{R}einforced \textbf{F}low \textbf{M}atching (\textbf{ARFM}), a novel adaptive offline RL fine-tuning method for  VLA flow models.
We control the strength of the RL signal introduced into the sample data by introducing an adaptively adjusted scaling factor in the VLA flow loss function. 
Specifically, we theoretically construct a variance-bias trade-off optimization objective with respect to the scaling factor. 
We aim to dynamically select the scaling factor to preserve the RL signal, i.e., samples with higher RL advantage are still amplified in the data distribution during flow model fine-tuning. 
Meanwhile, we hope to control the flow matching loss gradient variance, i.e., avoid some weights from exponentially exploding, resulting in excessive gradient variance and training crash.

Therefore, we focus on adaptively balancing “retaining enough RL advantage signal” and “controlling loss gradient variance” in the scaling factor as the distribution of the current batch data changes during the post-training process of the VLA flow model.
We theoretically construct an optimization objective for the scaling factor and construct an approximately analyzable objective function through some reasonable assumptions. 
A bisection iterative algorithm for real-time updating of the scaling factor is further induced, so that the loss of the VLA flow model is adaptively adjusted and corrected according to the data distribution.
Our principled contributions include three aspects:
\begin{itemize} 
\item We propose ARFM, a novel offline RL post-training method for VLA flow models that can adaptively adjust the data quality distribution. 
\item We theoretically establish the optimization objective of adaptively adjusting the scaling factor and induce a bisection iterative algorithm to update the factor in real time, thereby achieving efficient VLA flow model fine-tuning.
\item Extensive experiments in simulation and real-world robot manipulation tasks show that our proposed ARFM exhibits state-of-the-art generalization ability, robustness to dynamic perturbations, and excellent performance in few-shot and continuous learning scenarios.
\end{itemize}

\section{Related Work}
\label{sec:related_work}
\textbf{Flow matching models in RL.} Flow matching (\citealp{FM}) is a more general counterpart to the diffusion model (\citealp{DBLP:journals/corr/abs-2006-11239}).
Both diffusion and flow matching tend to have good results in offline RL. 
Some previous work explored modeling behavior policies based on diffusion models~\cite{janner2022planning,wangdiffusion}.
Following these studies, some researchers modeled offline RL objectives as energy-guided diffusion processes~\cite{chenoffline,lu2023contrastive}, while others adopted the same policy optimization approach utilizing diffusion and flow matching models without classifiers~\cite{ajayconditional,zheng2023guided}. 
Recently, an energy-weighted flow matching method~\cite{EWFM} for learning energy-guided distributions was proposed, which can directly utilize RL rewards or other metrics as energy functions without learning intermediate energy models.
Compared with previous studies, our work further extends the energy-weighted flow matching method to the VLA model post-training setting and proposes a new adaptive energy-weighted algorithm.

\textbf{RL for VLA models.} RL has recently emerged as a pivotal technique for post-training VLA models, overcoming the limitations of vanilla imitation learning~\cite{zhai2024fine,mark2024policy} on infrastructure such as RT-1~\cite{rt_1}.
These efforts are mainly divided into online and offline RL fine-tuning.
Online methods exploit direct environment interaction, adopt algorithms such as PPO (\citealp{schulman2017proximal}) or develop more data-efficient interaction frameworks~\cite{lu2025vla,tan2025interactive,guo2025improving}.
On the other hand, offline methods learn from static datasets by introducing various signals, such as human preferences~\cite{chen2025fdpp,zhang2024grape} and value guidance~\cite{nakamoto2024steering}.
Recent work~\cite{reinbot} has effectively implemented the core idea of RL in predicting the future to maximize return when examining offline RL post-training of VLA models.
The idea of utilizing RL returns as fine-grained goals to guide the policy comes from the previous new paradigms of sequence modeling, Decision Transformer~\cite{chen2021decision} and Reinformer~\cite{zhuang2024reinformer}.
Different from these studies, our work considers the implementation of offline RL post-training in the energy-weighted VLA flow model to achieve efficient and stable policy optimization.

\section{Preliminaries}
In this section, we first list the energy-weighted flow matching and give two equivalent loss functions to train a neural network policy to approximate the energy-guided flow. 
We then list the RL advantage evaluation method for post-training data and the expressive VLA flow model.
\subsection{Energy-weighted flow matching}
\begin{definition}
\label{def:1}
    Both flow matching (\citealp{FM}) and Energy-Weighted Flow Matching (EWFM)~\cite{EWFM} learn the \textit{vector field} \cite{chen2019neuralordinarydifferentialequations}. Consider a probability density path: $p: [0,1]\times \mathbb{R}^d\to \mathbb{R}_{\geq 0}$, $\xb\mapsto p_t(\xb)$. Define the  \textit{vector field} $\vb: [0,1]\times \mathbb{R}^d \to \mathbb{R}^d$ which \textit{generates} the probability density path $p$, satisfying: 
\begin{align}
\label{equa:def}
    \frac{d}{dt}p_t(\xb) + \mbox{div}(\vb_t(\xb)p_t(\xb))= 0.
\end{align}
\end{definition}
For the conditional distribution $\ptzero$, the conditional vector field is defined in the same way.
Consider a probability distribution $p_0(\xb_0)$ and an energy function $\cE(\cdot)$, then the energy-guided distribution satisfies: 
\begin{align}
\label{equa:1}
    q_0(\xb_0)\propto p_0(\xb_0)\exp(-\beta\cE(\xb_0)).
\end{align}
Then consider adding Gaussian noises as conditional distributions to the two distributions: $\ptzero = \qtzero:= \mathcal{N}(\xb_t|\alpha_t\xb_0,\sigma_t^2\mathbf{I})$. Define  $\pt:=\int_{\xb_0}\ptzero\pzero\mathrm d\xb_0$, $\qt := \int_{\xb_0}\qtzero\qzero \mathrm d\xb_0$ as the marginal distributions at time $t$, then energy-weighted flow matching aims to efficiently learn the vector field $\uthat$ which generates $\qt$.
\textbf{Theorem}~\ref{thm:1} calculates the vector field $\uthat$ which generates $q$, and \textbf{Theorem}~\ref{thm:2} finds the equivalent loss between learning the conditional vector field and the marginal vector field.
The proof process is in the Appendix.
\begin{theorem}
\label{thm:1}
    \textit{
    Suppose $p_0(\xb_0)$, $q_0(\xb_0)$ are defined in Eq.~\eqref{equa:1}, and the conditional distributions $\ptzero,\qtzero$, marginal distributions $\pt,\qt$ are also defined above. Consider the conditional vector field $\utzero$ which generates $\ptzero$, and the vector field $\uthat$ which generates $\qt$, then we have:\\
    \begin{align}
        \uthat = \int_{\xb_0} p_{0t}(\xb_0 | \xb) \utzero \frac{\exp(-\beta \cE(\xb_0))}{\exp(-\cE_t(\xb))}\mathrm d\xb_0\label{eq:energy-guided-flow},
    \end{align}
    where $\cE_t(\xb)$ is an intermediate energy function which is defined as:
    $$\cE_t(\xb) = -\log \mathbb{E}_{p_{0t}(\xb_0|\xb)}[\exp(-\beta \cE(\xb_0))].$$
    }
\end{theorem}
\textbf{Theorem}~\ref{thm:1} finds out a closed form of the vector field $\ut$, but it still remains challenging to learn $\ut$, since the intermediate energy function is unknown during the process, therefore \textbf{Theorem}~\ref{thm:2} gives the method to simplify the training process.
\begin{theorem}
\label{thm:2}
\textit{
$p_0(\xb_0),q_0(\xb_0),\pt,\qt,\cE(\xb_0),\cE_t(\xb_t),\\ \uthat,\utzero$ are all defined above. 
Consider learning a model $\vb_{\theta}(\xb)$ with the learnable parameter $\theta$, then we define the Energy-weighted Flow Matching (EFM) loss $\mathcal{L}_{EFM}$ as: 
$$\mathcal{L}_{EFM}(\theta)=\mathbb{E}_{\xb,t}[\frac{\exp(- \cE_t(\xb))}{\mathbb{E}_{\tilde{x}\sim p_t(\tilde \xb)}[\exp(- \cE_t(\tilde \xb))]}||\vb_{\theta}(\xb)-\uthat||^2],$$
and the Conditional Energy-weighted Flow Matching (CEFM) loss  $\mathcal{L}_{CEFM}$ as:
\begin{align*}&\mathcal{L}_{CEFM}(\theta)=\\
&\mathbb{E}_{\xb,t}[\frac{\exp(-\beta\cE(\xb_0))}{\mathbb{E}_{\tilde{x}\sim p_0(\tilde \xb_0)}[\exp(-\beta\cE(\tilde \xb_0))]}||\vb_{\theta}(\xb)-\utzero||^2],\end{align*}
where the expectation on $t$ is taken over some predefined distribution $\lambda(t)$, $\xb_0$ is sampled from the data distribution $p_0(\cdot)$ and $\xb$ at time $t$ is sampled by $\pt$ with conditional distribution $\ptzero$ generated by the flow $\utzero$. Then the two losses are equal up to a constant factor to $\theta$. Hence, $$\nabla_{\theta}\mathcal{L}_{EFM}(\theta) = \nabla_{\theta}\mathcal{L}_{CEFM}(\theta).$$
}

\end{theorem}
\textbf{Theorem}~\ref{thm:2} proves that the training process of the vector field of the marginal energy-guided distribution could be simplified to learning the conditional vector field, without the calculation of the intermediate energy function.
In the VLA offline RL post-training process, given the observation and action actor $\ob_t,\Ab_t$, and compared with the vanilla imitation learning action distribution $p(\Ab_t|\ob_t)$, we consider designing a new energy function $\cE(\Ab_t,\ob_t)$ and further extend the EWFM method to fine-tune the energy-guided VLA model distribution $\pi(\Ab_t|\ob_t)\propto p(\Ab_t|\ob_t)\exp(\beta \cE(\Ab_t,\ob_t))$.

\subsection{RL advantage signal estimation}
Leave-one-out~\cite{kool2019buy} estimates the RL advantage using sampling. 
The REINFORCE Leave-One-Out generates  $ K $  independent samples  $ x_1, \dots, x_K \sim p_{\theta}(\cdot|c) $  and utilizes all other samples to compute a baseline for the current return:
$ R^*(c, x_k) = R(c, x_k) - \frac{1}{K-1} \sum_{i=1, i \neq k}^{K} R(c, x_i) $. 
The equivalent form of this objective is:
\begin{equation}
    R^*(c, x_k) = \frac{K}{K - 1} \left( R(c, x_k) - \frac{1}{K} \sum_{i=1}^{K} R(c, x_i) \right).
\label{eq:RL_adv}
\end{equation}
This is a simple, unbiased, and low-variance advantage estimate, so we are inspired by it to design a critic-free offline RL advantage signal for the VLA multi-task setting.

\subsection{VLA flow model}
The $\pi_0$ model~\cite{pi_0} is a recently proposed VLA flow model with excellent performance, which mainly consists of a language model transformer backbone.
Following the standard late fusion VLM recipe, image encoders embed the robot's image observations into the same embedding space as language tokens. 
The observation $\ob_t$ consists of multiple RGB images, a language command, and the robot's proprioceptive state. Formally, $\ob_t = [I_1^t, \dots, I_n^t, \ell^t, q^t]$, where $I_i^t$ is the $i$-th image, $\ell^t$ is a sequence of language tokens, and $q^t$ is a vector of joint angles.
The action chunk $\Ab_t = [\ab_t, \ab_{t+1}, \dots, \ab_{t+H-1}]$ corresponds to a sequence of future actions, where $H$ is the action horizon.
The Flow Matching (FM) loss is given by:
\begin{equation}
    L_{FM}(\theta) = \mathbb{E}_{p(\Ab_t|\ob_t), q(\Ab_\tau^t|\Ab_t)} \left[ \left\| \vb_\theta(\Ab_\tau^t, \ob_t) - \ub(\Ab_\tau^t|\Ab_t) \right\|^2 \right],
\end{equation}
where $\vb_\theta(\Ab_\tau^t, \ob_t)$ is the model's output, $\ub(\Ab_\tau^t|\Ab_t)$ is the denoising vector field, and $q(\Ab_\tau^t|\Ab_t) = \mathcal{N}(\tau \Ab_t, (1 - \tau)I)$ is the noise distribution.

The noisy actions $\Ab_\tau^t$ are computed as $\Ab_\tau^t = \tau \Ab_t + (1 - \tau)\epsilon$, where $\epsilon \sim \mathcal{N}(0, I)$.
During inference, actions are generated by integrating the learned vector field from $\tau = 0$ to $\tau = 1$, starting with random noise $\Ab_0^t \sim \mathcal{N}(0, I)$. 
The integration rule is given by:
    \begin{equation}
        \Ab_{\tau+\delta}^t = \Ab_\tau^t + \delta \vb_\theta(\Ab_\tau^t, \ob_t),
    \end{equation}
where $\delta$ is the integration step size. 
It is worth noting that our adaptive offline RL post-training theoretical analysis and the induced fine-tuning algorithm can be applied but not limited to the $\pi_0$ model.

\begin{figure}[tbp]
\centering
\includegraphics[width=1\columnwidth]{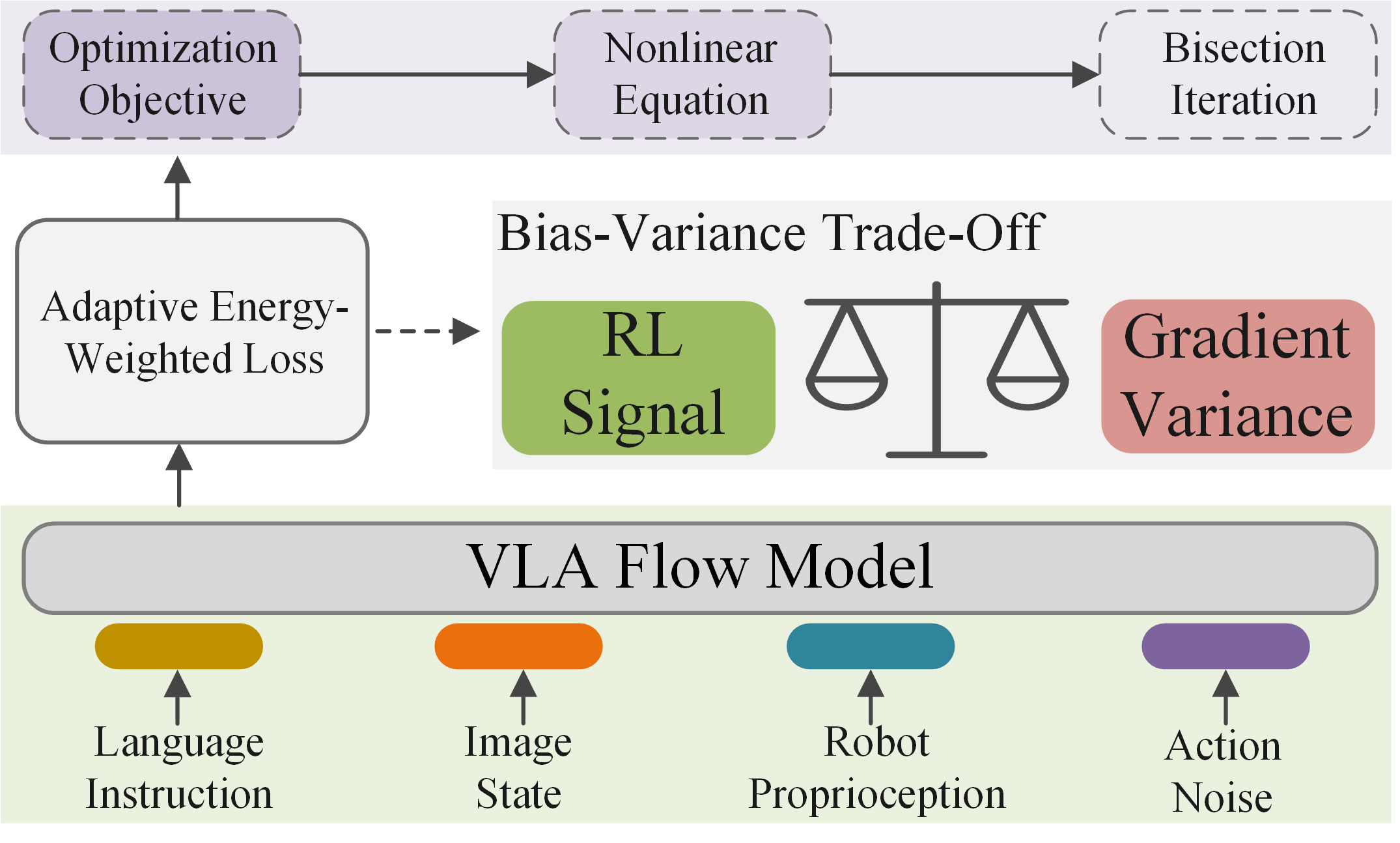}
\caption{The proposed ARFM method.
We study the offline RL post-training process of the VLA flow model, where the model input includes language instructions, external image perception and proprioception, and action noise that needs to be restored. We theoretically construct an energy-weighted flow matching loss with adaptive scaling factor, aiming to balance the RL signal and gradient variance on the data samples. Finally, we establish a solvable optimization objective and obtain a nonlinear equation about the scaling factor, which is finally solved by bisection iteration.
}
  \label{fig:pipline}
\end{figure}

\section{Methodology}
This work aims to construct a novel efficient post-training method for the VLA Flow model and induce a practical offline RL fine-tuning algorithm, as shown in Fig.~\ref{fig:pipline}. 
First, we construct an energy-weighted VLA model and the corresponding energy-weighted flow matching loss. 
Then, we explain in detail how to construct the optimization objective of the adaptive scaling factor in the loss function, which aims to weigh the RL signal and gradient variance of the data samples. 
Finally, we give the solution equations of the scaling factor under some reasonable assumptions, as well as the corresponding bisection iteration and the fine-tuning algorithm of the VLA flow model.

\subsection{Energy-weighted VLA flow model}
Consider a data distribution $p(\Ab_t|\ob_t)$, where $\Ab_t=[\ab_{t},\dots \ab_{t+H}]$ is the action chunk, and $\ob_t$ is the observation containing images in different views, language tokens, and  a vector of joint angles. The aim is to sample from the energy-guided distribution, which is: $$\pi(\Ab_t|\ob_t)\propto p(\Ab_t|\ob_t)\exp(\alpha R^{*}(\ob_t,\Ab_t)).$$
Here $\alpha $ is scaling factor, and $R^*(\ob_t,\Ab_t)$ is the return-to-go advantage. We could learn the vector field of the policy distribution $\pi$ by optimizing CEFM loss: 
$$L^{\tau}(\theta) = \mathbb{E}[\cE^*(\Ab_t,\ob_t)||\vb_{\theta}(\Ab^{\tau}_t,\ob_t)-\ub(\Ab^{\tau}_t|\Ab_t)||^2],$$
$$\mbox{where } \cE^*(\Ab_t,\ob_t)=\frac{\exp(\alpha R^*(\Ab_t,\ob_t))}{\mathbb{E}_{\Ab_t^*\sim p(\cdot|\ob_t)}\exp(\alpha R^*(\Ab_t^*,\ob_t))}.$$
Here in the expectation in $L^{\tau}(\theta)$, the action $\Ab_t$ is sampled from the distribution $p(\cdot|\ob_t)$, while $\Ab_t^{\tau}$ is to add a simple linear Gaussian noise (Optimal Transport) on $\Ab_t$, which is $\Ab_t^{\tau} = \tau \Ab_t+(1-\tau)\mathbf{\epsilon}$, where $\epsilon \sim \mathcal{N}(0,1)$ is a standard Gaussian variable. $\ub(\Ab_t^{\tau}|\Ab_t) $ is the conditional vector field, which could be formulated by: $\mathbf{u}(\Ab_t^{\tau}|\Ab_t) = \epsilon - \Ab_t$.

In practical post-training, we utilize the sampling assumption instead of directly calculating the expectation. 
In each step, we sample a batch of $B$ data pairs of $(\ob_t,\Ab_t)$, and calculate the softmax of the energy function as the energy weight $\cE^*$, and then calculate the weighted loss $L^{\tau}(\theta)$ within this batch, so the practical loss function should be:$$L^{\tau}_1(\theta) = \sum_{i=1}^{B}w_i(\alpha)||\vb_{\theta}(\{\Ab_t^i\}^{\tau},\ob_t) - \ub(\{\Ab_t^i\}^{\tau}|\Ab_t^i)||^2,$$
$$\mbox{where } w_i(\alpha)=\frac{ \exp(\alpha R^{*}(\Ab_t^i,\ob_t))}{\sum_{j=1}^{B}\exp(\alpha R^{*}(\Ab_t^j,\ob_t))}.$$ 
Here $\{\Ab_t^{i}\}$ are a batch sampled from $p(\cdot|\ob_t)$, while $\{\Ab_t^{i}\}^{\tau}$ is the noisy action from $\Ab_t^i$.
We utilize the standardized return $R^{*}$ as the energy weight in the loss function $L^{\tau}_1(\theta)$. 
The standardization here is performed according to the same task type in the VLA multi-task setting to facilitate the comparison of RL advantages of different states (by utilizing Eq.~\ref{eq:RL_adv}). Moreover, $\alpha$ in the loss function $L^{\tau}_1(\theta)$ is an important adaptive parameter, and we adaptively adjust its value by constructing an optimization objective that trades off between RL signal and gradient variance.

\subsection{Adaptive adjustment of the scaling factor $\alpha$}
The scaling factor $\alpha$ plays an important role in post-training of the VLA flow model.
Intuitively, if $\alpha$ is too small, the advantage of the data sample may not be fully reflected, so the improvement on the original flow matching is not significant. 
When $\alpha=0$, energy-weighted fine-tuning degenerates into vanilla flow matching. 
When $\alpha$ is large, fine-tuning tends to focus only on those data samples with higher energy. 
This will cause the sample data variance to be too large, resulting in gradient explosion and thus destroying the stability of fine-tuning.

Therefore, the scaling factor $\alpha$ is adaptively adjusted in each fine-tuning step, and its adjustment direction is to minimize $J(\alpha)$:
$$J(\alpha) = \mbox{Var}(\hat{g}(\alpha))-\lambda S(\alpha),$$
where $\hat{g}(\alpha)=\nabla_{\theta}L_1^{\tau}(\theta)=\sum_{i=1}^B\hat{w}_i(\alpha)\nabla_{\theta}||\vb_{\theta}(\{\Ab_t^i\}^{\tau},\ob_t) - \ub(\{\Ab_t^i\}^{\tau}|\Ab_t^i)||^2$ is the gradient of the loss $L_1^{\tau}(\theta)$, so $\alpha$ tends to minimize the variation of the gradient. $S(\alpha)=\sum_{i=1}^{B}\hat{w}_i(\alpha)R^*(\Ab_t^{i},\ob_t)/\sum_{i=1}^{B}\hat{w}_i(\alpha)$ is a score function which shows the effect of the RL advantage as the energy weight, and $\alpha$ tends to maximize the score.  $\hat{w}_i=\exp(\alpha R^{*}(\Ab_t^i,\ob_t))$ is the energy weight. $\lambda$ is a hyperparameter, to adjust the ratio between the RL signal and the gradient variance.

\begin{algorithm}[tb]
\caption{Bisection Iteration of Scaling Factor $\alpha$}
\label{alg:alpha}
\textbf{Input:} RL advantages $R^*_i$, FM losses $L_{FM}^i$, batch size $B$, hyperparameter $\lambda$, value range of $\alpha$ $[\alpha_{min},\alpha_{max}]$, number of iterations $M$ and tolerance $\epsilon$. 
\begin{algorithmic}[1]
\STATE Calculate $\sigma_R^2=\sum_iR_i^2/B$, $\mu_L$ =$\sum_i L^i_{FM}/B$, $\sigma_L^2 = \sum_i(L_{FM}^i - \mu_L)^2/B$, $x_{low} = \sigma_A^2\alpha_{min}$, $x_{high} = \sigma_A^2\alpha_{max}$
\STATE Define $F(x) = 4\sqrt{x}e^{2x}-2\sqrt{x}e^{x}-\frac{\lambda\sigma_R}{\sigma_L^2}$
\FOR{$m=1$ to $M$}
\STATE$x_{mid} = 0.5(x_{low}+x_{high})$
\STATE \textbf{if} {$|F(x_{mid})|<\epsilon$}: \textbf{Break}
\STATE \textbf{if} {$F(x_{mid})>0$}: $x_{high} = x_{mid}$ 
\STATE \textbf{else}: $x_{low} = x_{mid}$
\ENDFOR
\STATE$\alpha^* = \sqrt{0.5(x_{low}+x_{high})}/\sigma_A$
\STATE \textbf{Return} clip ($\alpha^*,\alpha_{max},\alpha_{min}$)
\end{algorithmic}
\end{algorithm}
\begin{algorithm}[tb]
\caption{ARFM: Post-Training of VLA Flow Model}
\label{alg:Post-training of VLA Flow model}
\textbf{Input}: Post-traing data $\{\Ab_t,\ob_t\}$, batch size $B$, VLA flow model $\vb_{\theta}$.
\begin{algorithmic}[1] 
\FOR{a batch of data $\{\Ab_t^i,\ob_t^i\}$}
\FOR{$i$ in $[B]$}
\STATE Sample $\epsilon_i\sim\mathcal{N}(0,\Ib)$, $\tau\sim \mbox{Uniform}(0,1)$
\STATE $\{\Ab_t^i\}^\tau = \tau(\Ab_t^i) + (1-\tau) \epsilon_i$
\STATE $R^i=R^*(\Ab_t^i,\ob_t^i)$, $g_i = \exp(R^*(\Ab_t^i,\ob_t^i))$
\STATE $L_{FM}^i = ||\vb_{\theta}(\{\Ab_t^{i}\}^\tau,\ob_t)-(\epsilon - \Ab_t^i)||^2$
\ENDFOR
\STATE Calculate the optimal $\alpha^*$ from Alg.~\ref{alg:alpha}
\STATE $w_i(\alpha^*) = \exp(\alpha^* g_i)/\sum_j\exp(\alpha^* g_j)$
\STATE $L_1^\tau(\theta) = \sum_i w_i(\alpha^*)L_{FM}^i$
\STATE Take a gradient step of $L_1^\tau(\theta)$
\ENDFOR
\end{algorithmic}
\end{algorithm}

Intuitively, minimizing $J(\alpha)$ is to balance the trade-off to emphasize the energy weight and prevent the post-training from gradient explosion.
In order to make this optimization objective solvable, three assumptions need to be considered:
\begin{assumption}
\textit{
The standardized RL advantage signal $R^*(\Ab_t,\ob_t)$ is assumed to be Gaussian variable $\mathcal{N}(0,\sigma_R^2)$.
}
\end{assumption}
\begin{assumption}
\textit{
The Conditional Flow Matching (CFM) loss $L_{CFM}^i=||\vb_{\theta}(\Ab_t^i,\ob_t^i)-\ub(\{\Ab_t^i\}^\tau|\Ab_t^i)||^2$ is also assumed to be Gaussian variable $\mathcal{N}(\mu_L,\sigma_L^2)$.    
}
\end{assumption}
\begin{assumption}
\textit{
When the batch size $B$ is large enough, the sample-based expectation and variance could be utilized to approximate $\alpha_L,\sigma_A,\sigma_L$. 
}
\end{assumption}
In the post-training stage of the VLA flow model, the RL advantage is standardized, and the CFM loss value will quickly tend to have a lower variance, so the Gaussian distribution assumption here is mild and reasonable.
Based on these assumptions, we can obtain two important corollaries on the solution of the scaling factor $\alpha$. 
The derivation details are in the Appendix.
\begin{corollary}
    \label{cor:alpha1}
    \textit{
    With the assumptions and functions defined above, the  objective $J(\alpha)$ could be represented by:
    \begin{align}J(\alpha) =\sigma_L^2[e^{2\alpha^2\sigma_R^2}-e^{\alpha^2\sigma_R^2}]-\lambda\alpha\sigma_R^2.\end{align}
    }
\end{corollary}
\begin{corollary}
    \label{cor:alpha2}
    \textit{
    The $\alpha^*$ which minimizes $J(\alpha)$ is solved by:\\
    \begin{align}
       4\sqrt{x^*}e^{2x^*}-2\sqrt{x^*}e^{x^*}-\frac{\lambda\sigma_R}{\sigma_L^2} = 0, \alpha^*=\frac{\sqrt{x^*}}{\sigma_R}.
    \end{align}
    }
\end{corollary}
\textbf{Corollary}~\ref{cor:alpha1} gives a specific solvable optimization objective for $\alpha$. 
\textbf{Corollary}~\ref{cor:alpha2} gives the corresponding nonlinear equation for $\alpha$, which can be quickly solved utilizing bisection iteration (Alg.~\ref{alg:alpha}). 
Therefore, we can utilize Alg.~\ref{alg:alpha} to obtain the optimal $\alpha^*$, thereby obtaining the loss based on energy-weighted flow matching. 
This adaptive loss function is finally utilized in the post-training process of the VLA flow model (Alg.~\ref{alg:Post-training of VLA Flow model}).

\section{Experiments}
In this section, 
we conduct extensive experiments in various scenarios to evaluate the effectiveness of the proposed ARFM method.
Specifically, we aim to examine the following five questions:
\textbf{1)} Does ARFM show superior generalization ability compared to previous SOTA baseline methods?
\textbf{2)} How resistant is ARFM to action noise in robot manipulation tasks with action noise?
\textbf{3)} How does ARFM perform in data-scarce scenarios and lifelong learning, especially in terms of few-shot learning and continuous learning capabilities?
\textbf{4)} To what extent do the key hyper-parameters in the ARFM method affect the performance of the VLA flow model?
\textbf{5)} How does ARFM perform in real-world robot manipulation tasks, especially manipulating disturbed objects?

\begin{figure}[ht]
\centering
\includegraphics[width=1\columnwidth]{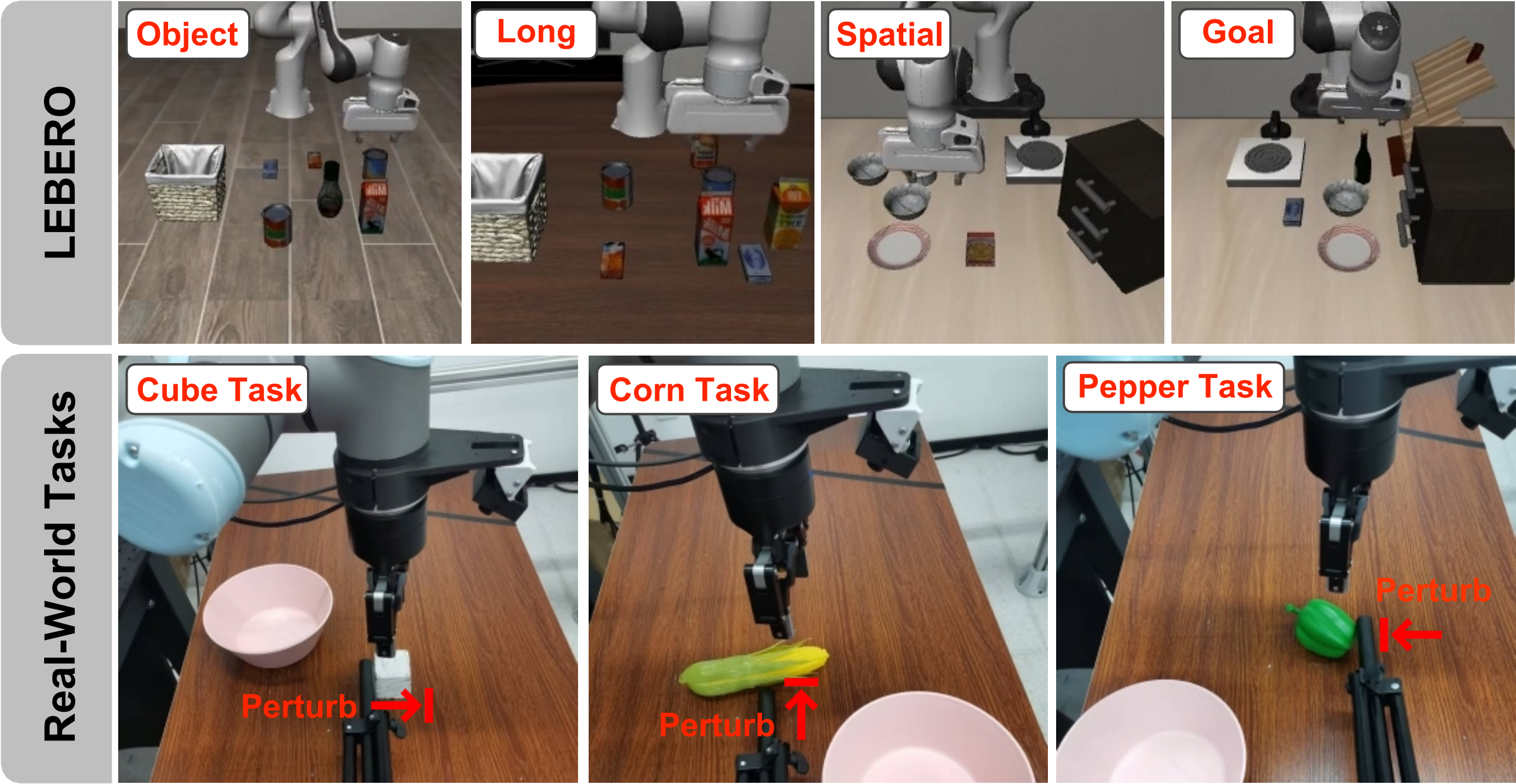}
\caption{We evaluate performance on four categories of the LIBERO benchmark suite (Object, Long, Spatial and Goal) and three categories of real-world UR5 grasping and placing tasks. 
In realistic experiments, we introduce multiple external perturbations to the objects to be manipulated.
}
  \label{fig:exp_setting_1}
\end{figure}

\begin{table*}[!ht]
    \begin{center}
            \renewcommand\arraystretch{1.05}
             \setlength\tabcolsep{15pt}
    \begin{tabular}{c|c|cccc|c}
    \hline \thickhline
    &  & \multicolumn{5}{c}{\textbf{LIBERO Multi-task Learning}}\\
    \cline{3-7}
    \multicolumn{1}{c|}{\multirow{-2}{*}{\textbf{Model Type}}} &\multicolumn{1}{c|}{\multirow{-2}{*}{\textbf{Models}}}         &\textbf{Goal} &\textbf{Spatial} &\textbf{Object} &\textbf{Long} &\textbf{Average}\\
    \hline
    \hline
    \multirow{6}{*}{{Non-Flow Matching}}
     &Octo                    &84.6  &78.9  &85.7  &51.1 &75.1  \\ 
     &OpenVLA                 &79.2  &84.7  &88.4  &53.7 &76.5  \\
     &Dita                    &85.4  &84.2  &\textbf{96.3}  &63.8 &82.4  \\
     &Diffusion Policy              &68.3  &78.3  &92.5  &50.5 &72.4  \\
     &MDT                           &73.5  &78.5  &87.5  &64.8 &76.1  \\
     &QueST                         &80.8 &87.4 &93.6 &68.8 &82.7  \\ 
     \hline
    \multirow{4}{*}{{Flow Matching}}
     &$\pi_0$                       &93.8 &91.2 &93.2 &74.2 &88.1  \\
     &ReinboT                       &94.0 &95.6 &93.8 &81.4 &91.2$_{{(+3.5\%)}}$  \\
     &RWR                           &94.4 &94.0 &94.3 &80.4 &90.8$_{{(+3.1\%)}}$  \\
     &\textbf{ARFM (Ours)}                          &\textbf{94.9} &\textbf{95.8} &95.0 &\textbf{82.6} &\textbf{92.1$_{{\mathbf{(+4.5\%)}}}$}  \\
     
     \hline \thickhline
    \end{tabular}
    \end{center}
    \caption{Multi-task Success Rate (SR) on the LIBERO benchmark, with the best results highlighted in bold.}
    \label{table:comparison}
\end{table*}

\textbf{Experimental setup.}
We evaluate the performance of the proposed ARFM method in LIBERO~\cite{liu2023libero} simulation and the real-world UR5 platform (Fig.~\ref{fig:exp_setting_1}).
The LIBERO is a comprehensive lifelong learning benchmark that encompasses multiple task suites.
These suites are designed to assess specific aspects of general robot manipulation, with tasks defined through language-guided instructions. 
Specifically, LIBERO includes four main suites: Object, Long, Spatial and Goal. 
Each suite is tailored to test different object manipulation capabilities and comprises 10 distinct tasks. 
For real-world evaluation, we utilize a UR5 robotic arm to assess the ARFM's performance. 
We set up three distinct pick-and-place tasks to evaluate the model's robustness under physical perturbations.
Hyperparameters and reward configurations can be found in Appendix Tab.~\ref{tab:trainer} and Tab.~\ref{tab:rewards} respectively.

\textbf{Baselines.}
We mainly consider two types of baseline models $--$ non-flow matching type and flow matching type.
In the non-flow matching baseline model, we include the general auto-regressive models Octo~\cite{octo} and OpenVLA~\cite{openvla}.
We also consider diffusion-based models such as Diffusion Policy~\cite{chi2023diffusion},  MDT~\cite{reuss2024multimodal} and Dita~\cite{hou2025dita}.
In addition, we include QueST~\cite{mete2024quest}, which discretizes the continuous action space into a skill codebook through VQ-VAE and then predicts the skills auto-regressively.
On the other hand, for the flow matching type of baseline methods, we first consider $\pi_0$~\cite{pi_0}, which is a VLA flow model that utilizes trajectory-level flow matching to achieve efficient policy learning.
We also consider the offline RL fine-tuning baselines ReinboT~\cite{reinbot} and RWR~\cite{peters2007reinforcement}. 
ReinboT guides VLA action generation by predicting densely maximized future return, while RWR optimizes the VLA model by performing reward-weighted regression on samples. 
For a more fair comparison, we implemented the flow model versions of ReinboT and RWR based on the $\pi_0$ model.
The baseline reproduction details are in the Appendix.

\textbf{Multi-task learning setting.}
Tab.~\ref{table:comparison} compares ARFM with previous state-of-the-art baseline algorithms.
The results show that the performance of flow matching models is higher than that of non-flow matching models overall, which may be mainly attributed to the powerful trajectory modeling ability of flow matching models.
Moreover, among flow matching models, ARFM achieves the highest success rate $92.1\%$, which is $4.5\%$ higher than the baseline $\pi_0$. 
It is followed by ReinboT ($91.2$) and RWR ($90.8$) baseline algorithms, which are $3.5\%$ and $3.3\%$ higher than the baseline $\pi_0$, respectively.
The results preliminarily prove that our ARFM method can more efficiently fine-tune the VLA flow model through the adaptive energy-weighted scaling mechanism, thereby obtaining better generalization performance.

\textbf{Action perturbation setting.}
To evaluate the model's resistance to perturbations, we added different levels of Gaussian noise ($0.1\sim0.3$) to the actions inferred by the model during evaluation. 
The experimental results are shown in Tab.~\ref{table:action_noise}.
The results show that compared with the ReinboT ($46.3$) and RWR (46.4), ARFM has the highest average success rate ($48.2$), which is $11.4\%$ higher than the baseline $\pi_0$ ($43.3$).
This indicates that by balancing the RL signal and gradient variance, ARFM learns a more robust VLA model that is effectively resistant to action perturbations.

\begin{table}[t]
    
    \begin{center}
        \renewcommand\arraystretch{1.05}
        \resizebox{\linewidth}{!}{
        \begin{tabular}{c|cccc|c}
        \hline \thickhline
        & \multicolumn{4}{c|}{\textbf{LIBERO Action Perturbation}} \\ 
        \cline{2-5}
         \multicolumn{1}{c|}{\multirow{-2}{*}{\textbf{Models}}} & \textbf{Goal} & \textbf{Spatial} & \textbf{Object} & \textbf{Long} &\multicolumn{1}{c}{\multirow{-2}{*}{\textbf{Avg.}}} \\
        \hline \hline
        $\pi_0$          & 47.5 & 50.6 & 44.9 & 30.0 & 43.3 \\
        ReinboT          & \textbf{51.4} & 59.6 & 44.8 & 29.3 & 46.3$_{{(+6.9\%)}}$ \\
        RWR              & 49.5 & 60.1 & 46.9 & 29.1 & 46.4$_{{(+7.2\%)}}$ \\
        \textbf{ARFM (Ours)}    & 49.7 & \textbf{61.1} & \textbf{48.9} & \textbf{33.0} & \textbf{48.2$_{{\mathbf{(+11.4\%)}}}$} \\
        \hline \thickhline
        \end{tabular}
        }
    \end{center}
    \caption{Average SR(\%) under action perturbations (noise levels = 0.1, 0.15, 0.2, 0.25, 0.3) across four LIBERO suites. }
    \label{table:action_noise}
\end{table}

\begin{table}[t]
    
    \begin{center}
        \renewcommand\arraystretch{1.05}
        \resizebox{\linewidth}{!}{
        \begin{tabular}{c|ccc|c}
        \hline \thickhline
        & \multicolumn{3}{c|}{\textbf{LIBERO-Long Few-Shot}} \\ 
        \cline{2-4}
         \multicolumn{1}{c|}{\multirow{-2}{*}{\textbf{Models}}} & \textbf{30-shot} & \textbf{20-shot} & \textbf{10-shot} &\multicolumn{1}{c}{\multirow{-2}{*}{\textbf{Avg.}}} \\
        \hline \hline
        $\pi_0$          & 41.7 & 33.8 & 22.1 & 32.5 \\
        ReinboT              & 39.5 & 37.5 & 24.6 & 33.9$_{{(+4.1\%)}}$ \\
        RWR          & 39.5 & 37.7 & 26.7 & 34.6$_{{(+6.5\%)}}$  \\
        \textbf{ARFM (Ours)}    & \textbf{42.9} & \textbf{38.9} & \textbf{27.7} & \textbf{36.5$_{{\mathbf{(+12.2\%)}}}$} \\
        \hline \thickhline
        \end{tabular}
        }
    \end{center}
    \caption{Average SR(\%) of few-shot learning settings on LIBERO-Long task (noise levels = 0.1, 0.15, 0.20).}
    \label{table:few_shot}
\end{table}

\begin{figure}[t]
\centering
\includegraphics[width=0.8\columnwidth]{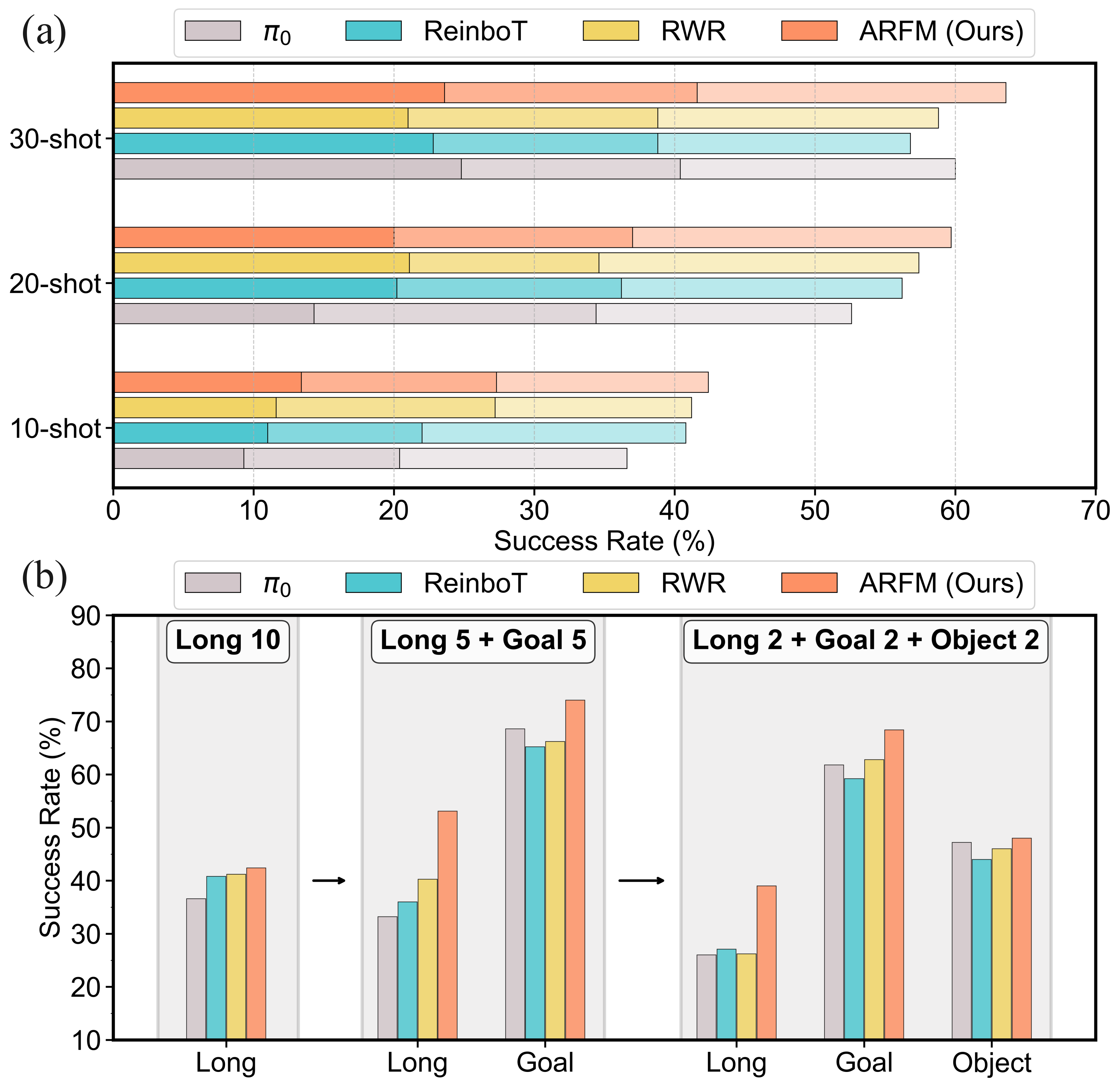}
\caption{(a) Average SR(\%) of different few-shot learning settings on LIBERO-Long under action perturbations (noise levels = 0.1, 0.15, 0.2). The three color shades from light to dark correspond to increasing levels of action noise (0.1 → 0.2). (b) Continuous learning on LIBERO benchmark (noise levels=0.1). Training sequence: Long 10 → Long 5 + Goal 5 → Long 2 + Goal 2 + Object 2. 
"Long 10" means that the model is trained on 10 trajectories per task on the LIBERO-Long suite, and other abbreviations are similar.
}
  \label{fig:few_shot_continuous_learning}
\end{figure}


\textbf{Few-shot learning setting.} We examine the few-shot learning capabilities of the model in the LIBERO-Long suite, as shown in Tab.~\ref{table:few_shot}, Fig. ~\ref{fig:few_shot_continuous_learning} (a) and Appendix Tab.~\ref{table:libero-few-shot-all}.
The experiment shows that ARFM performs best ($36.5$), followed by RWR ($34.6$) and ReinboT ($33.9$), and finally $\pi_0$ ($32.5$).
This shows that compared with the offline RL fine-tuning method introduced by the baseline ReinboT and RWR, ARFM's adaptive energy-weighted RL fine-tuning has better data utilization efficiency, thus reflecting better small sample learning and continuous learning performance.

\begin{table*}[!ht]

   
    \begin{center}
        \renewcommand\arraystretch{1.2}
        \resizebox{\linewidth}{!}{
        \begin{tabular}{c|cc|cc|ccc|c|c}
        \hline \thickhline
        \multirow{3}{*}{\textbf{Models}} & 
        \multicolumn{7}{c|}{\textbf{LIBERO Continual Learning}} & 
        \multirow{3}{*}{\centering{\textbf{Avg. NBT \(\downarrow\)}}} & 
        \multirow{3}{*}{\centering{\textbf{Avg. SR \(\uparrow\)}}} \\
        \cline{2-8}
        & \multicolumn{2}{c|}{\textbf{L 30 $\rightarrow$ L 15 + G 15}} & 
        \multicolumn{2}{c|}{\textbf{L 20 $\rightarrow$ L 10 + G 10}} & 
        \multicolumn{3}{c|}{\textbf{L 10 $\rightarrow$ L 5 + G 5 $\rightarrow$ L 2 + G 2 + O 2}} & & \\
        \cline{2-8}
        & \textbf{L NBT \(\downarrow\)} & \textbf{Avg. SR \(\uparrow\)} 
        & \textbf{L NBT \(\downarrow\)} & \textbf{Avg. SR \(\uparrow\)} 
        & \textbf{L NBT \(\downarrow\)} & \textbf{G NBT \(\downarrow\)} & \textbf{Avg. SR \(\uparrow\)} 
        & & \\
        \hline \hline
        $\pi_0$       & 3.2 & 64.3 & 9.4 & 55.8 & 10.6 & 6.8 & 45.6 & 7.5 & 55.2 \\
        ReinboT           & \textbf{0} & 63.1 & \textbf{6.7} & 59.2 & 13.7 & 6.0 & 45.4 & 6.6$_{{(-12.0\%)}}$ & 55.9$_{{(+1.2\%)}}$ \\
        RWR       & \textbf{0} & 60.3 & 11.4 & 58.6 & 14.5 & \textbf{3.4} & 47.1 & 7.3$_{{(-2.3\%)}}$ & 55.3$_{{(+0.2\%)}}$ \\
        \textbf{ARFM (Ours)} & 1.1 & \textbf{67.6} & 8.5 & \textbf{61.1} & \textbf{3.4} & 5.6 & \textbf{54.2} & \textbf{4.7$_{{\mathbf{(-38.0\%)}}}$} & \textbf{61.0$_{{\mathbf{(+10.5\%)}}}$} \\
        \hline \thickhline
        \end{tabular}
        }
    \end{center}
 \caption{
    Negative Backward Transfer (NBT) and Success Rate (SR) under continual learning on LIBERO-Long (L), Goal (G), and Object (O) suites (noise levels=0.1). 
    }
    \label{table:libero-nbt-sr}
\end{table*}

\textbf{Continual learning setting.} In the continual learning setting, we evaluate the model's ability to learn tasks sequentially without forgetting previously acquired skills. 
We utilize the Negative Backward Transfer (NBT) metric to measure how much the model's performance degrades on task $i$ after learning all tasks: $NBT = \frac{1}{T-1}\sum_{i}^{T-1}(max(0,(SR)_i-(SR)_i^T))$.
$(SR)_i$ indicates the success rate of the model on task $i$ after learning it, and $(SR)_i^T$ indicates the success rate of the model on task $i$ after learning all tasks.
The experiment result are in Tab.~\ref{table:libero-nbt-sr}, Fig.~\ref{fig:few_shot_continuous_learning} (b) and Appendix Tab.~\ref{table:libero-merged}.
Compared with the baseline $\pi_0$ (SR of $55.2$ and NBT of $7.5$), ARFM (SR of $61.0$ and NBT of $4.7$) not only improves the final average success rate by $10.5\%$, but also reduces the NBT by $38.0\%$.
This shows that ARFM is not only able to learn new tasks more effectively, but also significantly alleviates catastrophic forgetting, which is crucial for lifelong learning agents.

\subsection{Ablation study}

\begin{figure}[t]
\centering
\includegraphics[width=0.85\columnwidth]{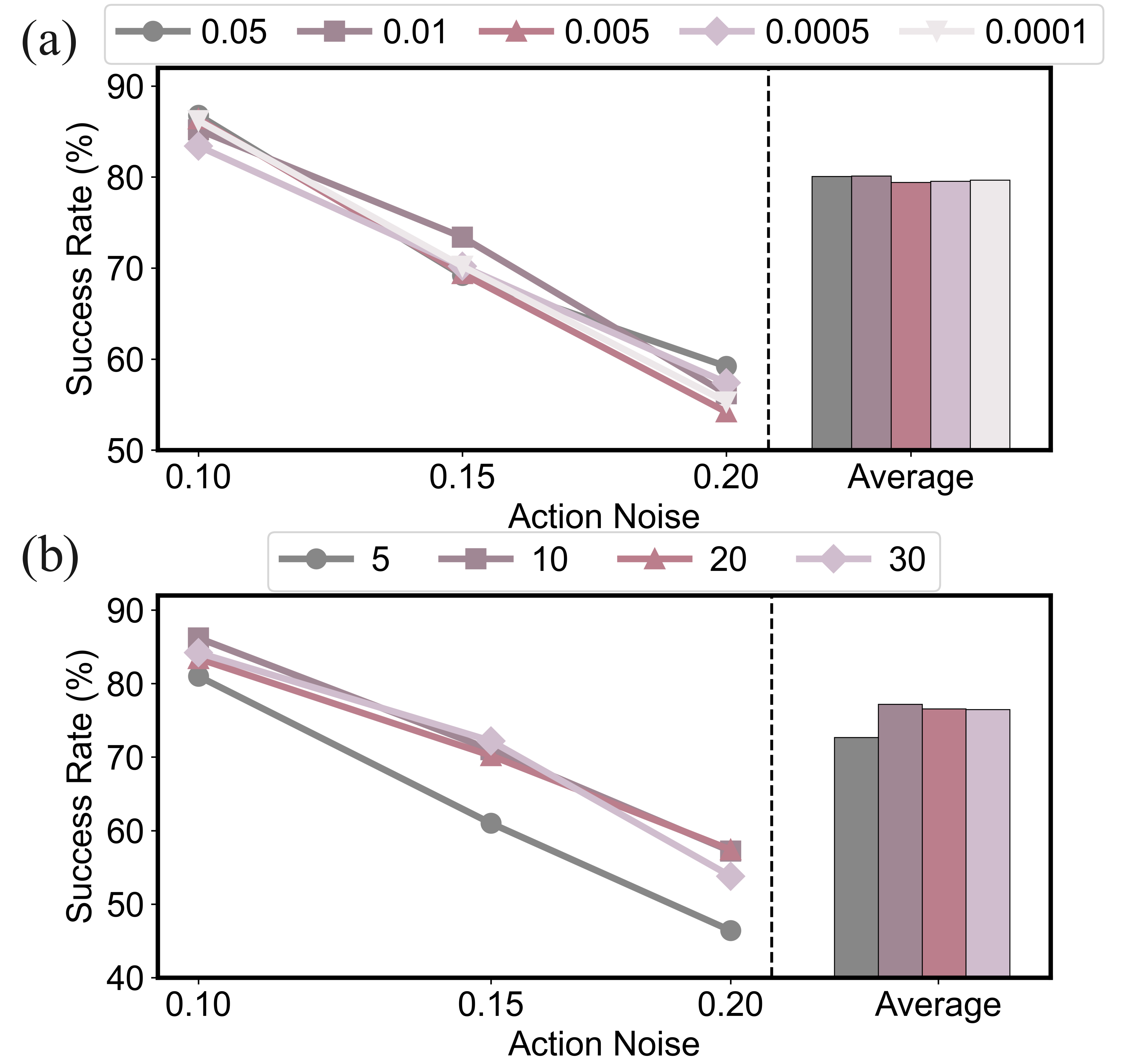}
\caption{
Ablation study in the LIBERO-Goal suite on the optimization objection hyperparameter $\lambda$ (a) and the number of bisection iterations $M$ (b).
}
  \label{fig:ablation_hyperparams}
\end{figure}

In the implementation of the proposed ARFM method, the optimization objective hyperparameter $\lambda$ and the total number of bisection iterations $M$ play an important role. 
The $\lambda$ is utilized to balance the relative magnitude between the RL signal and the gradient variance, while the $M$ directly affects the accuracy of the $\alpha^*$. 
The results in Fig.~\ref{fig:ablation_hyperparams} show that the performance of ARFM is insensitive to the $\lambda$, that is, different $\lambda$ values have little effect on the performance of the VLA flow model.
This may be mainly due to the adaptive adjustment ability of the loss weight of the ARFM method itself. 
In terms of the number of iterations $M$, the performance of the VLA flow model begins to stabilize when the $M$ value is $10$ or higher, indicating that ARFM can find an approximate optimal solution within a small number of steps.

\begin{figure}[t]
\centering
\includegraphics[width=0.85\columnwidth]{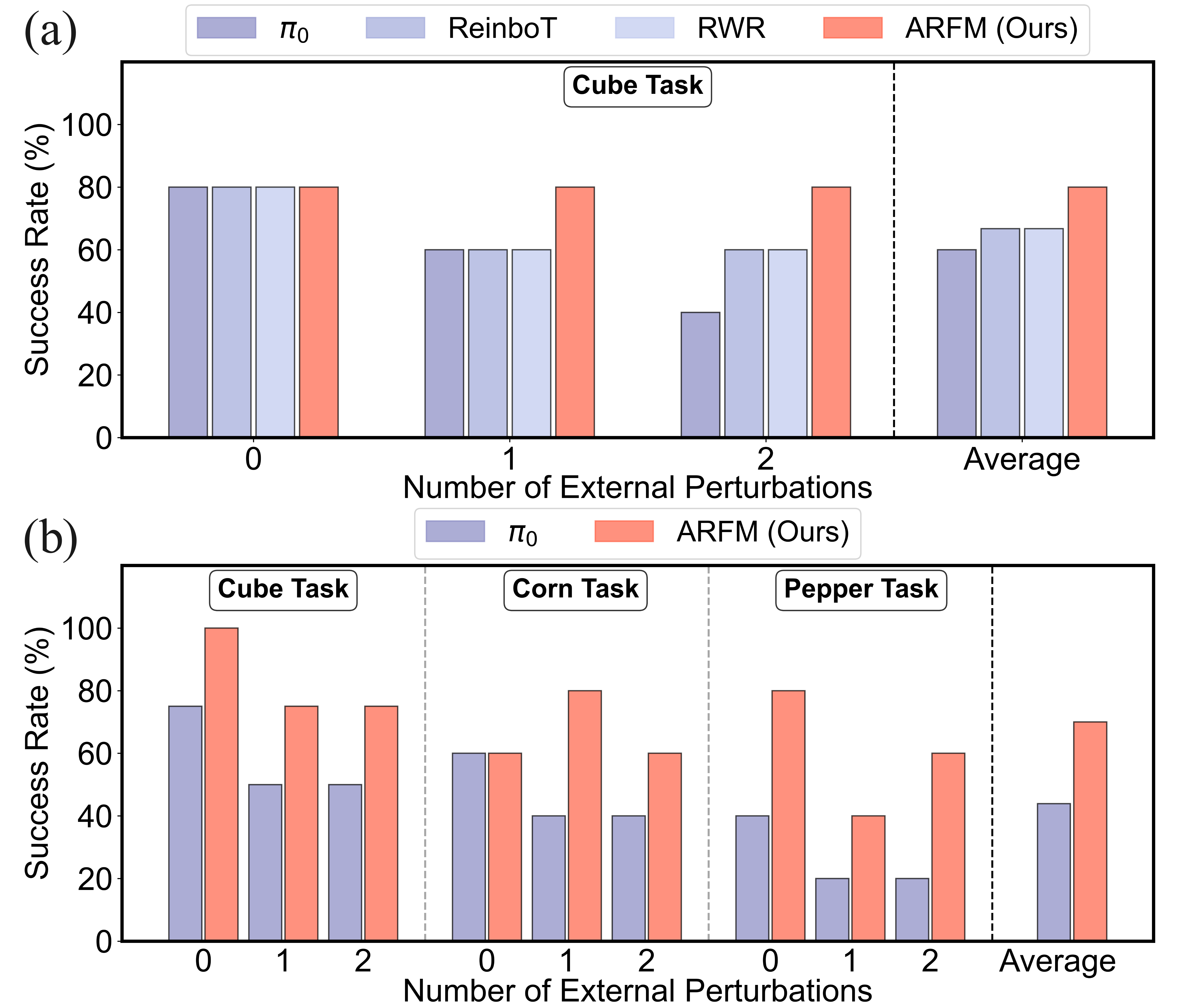}
\caption{
Performance comparison of real-world pick-and-place tasks under external perturbations.
}   
  \label{fig:real_world_2}
\end{figure}
\subsection{Real-world experiments}

We compare the performance of the models on some real-world grab and place tasks that are subject to external perturbations, as shown in Fig.~\ref{fig:real_world_2}. 
The experimental results show that the proposed ARFM achieves the best performance, followed by ReinboT and RWR, with $\pi_0$ being the worst. 
Moreover, the ability of ARFM to resist perturbation is significantly better than that of baseline $\pi_0$.
These results confirm that our adaptive offline RL fine-tuning method can enable the robot to perform more robust decision-making actions to cope with real-world complex scenarios.





\section{Conclusion}
We propose an adaptive offline RL post-training method for VLA flow models.
We consider the balance between retaining enough RL advantage signals and controlling the loss gradient variance, so that the loss can change adaptively according to the quality distribution of the current batch of post-training data.
Extensive experiments verify that ARFM has excellent generalization ability, robustness to dynamic perturbations, and few-shot learning and continuous learning capabilities.
A promising research direction in the future is to explore online RL post-training of VLA flow policies, that is, how to efficiently adapt to new scenarios by interacting with the environment.

\bibliography{aaai2026}

\begin{thebibliography}{41}
\providecommand{\natexlab}[1]{#1}

\bibitem[{Ajay et~al.(2022)Ajay, Du, Gupta, Tenenbaum, Jaakkola, and Agrawal}]{ajayconditional}
Ajay, A.; Du, Y.; Gupta, A.; Tenenbaum, J.~B.; Jaakkola, T.~S.; and Agrawal, P. 2022.
\newblock Is Conditional Generative Modeling all you need for Decision Making?
\newblock In \emph{The Eleventh International Conference on Learning Representations}.

\bibitem[{Alayrac et~al.(2022)Alayrac, Donahue, Luc, Miech, Barr, Hasson, Lenc, Mensch, Millican, Reynolds et~al.}]{alayrac2022flamingo}
Alayrac, J.-B.; Donahue, J.; Luc, P.; Miech, A.; Barr, I.; Hasson, Y.; Lenc, K.; Mensch, A.; Millican, K.; Reynolds, M.; et~al. 2022.
\newblock Flamingo: a visual language model for few-shot learning.
\newblock \emph{Advances in neural information processing systems}, 35: 23716--23736.

\bibitem[{Black et~al.(2024)Black, Brown, Driess, Esmail, Equi, Finn, Fusai, Groom, Hausman, Ichter et~al.}]{pi_0}
Black, K.; Brown, N.; Driess, D.; Esmail, A.; Equi, M.; Finn, C.; Fusai, N.; Groom, L.; Hausman, K.; Ichter, B.; et~al. 2024.
\newblock $\pi_0$: A Vision-Language-Action Flow Model for General Robot Control.
\newblock \emph{arXiv preprint arXiv:2410.24164}.

\bibitem[{Black et~al.(2023)Black, Nakamoto, Atreya, Walke, Finn, Kumar, and Levine}]{black2023zero}
Black, K.; Nakamoto, M.; Atreya, P.; Walke, H.; Finn, C.; Kumar, A.; and Levine, S. 2023.
\newblock Zero-shot robotic manipulation with pretrained image-editing diffusion models.
\newblock \emph{arXiv preprint arXiv:2310.10639}.

\bibitem[{Brohan et~al.(2023)Brohan, Brown, Carbajal, Chebotar, Chen, Choromanski, Ding, Driess, Dubey, Finn et~al.}]{rt_2}
Brohan, A.; Brown, N.; Carbajal, J.; Chebotar, Y.; Chen, X.; Choromanski, K.; Ding, T.; Driess, D.; Dubey, A.; Finn, C.; et~al. 2023.
\newblock Rt-2: Vision-language-action models transfer web knowledge to robotic control.
\newblock \emph{arXiv preprint arXiv:2307.15818}.

\bibitem[{Brohan et~al.(2022)Brohan, Brown, Carbajal, Chebotar, Dabis, Finn, Gopalakrishnan, Hausman, Herzog, Hsu et~al.}]{rt_1}
Brohan, A.; Brown, N.; Carbajal, J.; Chebotar, Y.; Dabis, J.; Finn, C.; Gopalakrishnan, K.; Hausman, K.; Herzog, A.; Hsu, J.; et~al. 2022.
\newblock Rt-1: Robotics transformer for real-world control at scale.
\newblock \emph{arXiv preprint arXiv:2212.06817}.

\bibitem[{Cadene et~al.(2024)Cadene, Alibert, Soare, Gallouedec, Zouitine, Palma, Kooijmans, Aractingi, Shukor, Aubakirova, Russi, Capuano, Pascale, Choghari, Moss, and Wolf}]{cadene2024lerobot}
Cadene, R.; Alibert, S.; Soare, A.; Gallouedec, Q.; Zouitine, A.; Palma, S.; Kooijmans, P.; Aractingi, M.; Shukor, M.; Aubakirova, D.; Russi, M.; Capuano, F.; Pascale, C.; Choghari, J.; Moss, J.; and Wolf, T. 2024.
\newblock LeRobot: State-of-the-art Machine Learning for Real-World Robotics in Pytorch.
\newblock \url{https://github.com/huggingface/lerobot}.

\bibitem[{Chen et~al.(2022)Chen, Lu, Ying, Su, and Zhu}]{chenoffline}
Chen, H.; Lu, C.; Ying, C.; Su, H.; and Zhu, J. 2022.
\newblock Offline Reinforcement Learning via High-Fidelity Generative Behavior Modeling.
\newblock In \emph{The Eleventh International Conference on Learning Representations}.

\bibitem[{Chen et~al.(2021)Chen, Lu, Rajeswaran, Lee, Grover, Laskin, Abbeel, Srinivas, and Mordatch}]{chen2021decision}
Chen, L.; Lu, K.; Rajeswaran, A.; Lee, K.; Grover, A.; Laskin, M.; Abbeel, P.; Srinivas, A.; and Mordatch, I. 2021.
\newblock Decision transformer: Reinforcement learning via sequence modeling.
\newblock \emph{Advances in neural information processing systems}, 34: 15084--15097.

\bibitem[{Chen et~al.(2019)Chen, Rubanova, Bettencourt, and Duvenaud}]{chen2019neuralordinarydifferentialequations}
Chen, R. T.~Q.; Rubanova, Y.; Bettencourt, J.; and Duvenaud, D. 2019.
\newblock Neural Ordinary Differential Equations.
\newblock arXiv:1806.07366.

\bibitem[{Chen et~al.(2025)Chen, Jha, Tomizuka, and Romeres}]{chen2025fdpp}
Chen, Y.; Jha, D.~K.; Tomizuka, M.; and Romeres, D. 2025.
\newblock FDPP: Fine-tune Diffusion Policy with Human Preference.
\newblock \emph{arXiv preprint arXiv:2501.08259}.

\bibitem[{Chi et~al.(2023)Chi, Xu, Feng, Cousineau, Du, Burchfiel, Tedrake, and Song}]{chi2023diffusion}
Chi, C.; Xu, Z.; Feng, S.; Cousineau, E.; Du, Y.; Burchfiel, B.; Tedrake, R.; and Song, S. 2023.
\newblock Diffusion policy: Visuomotor policy learning via action diffusion.
\newblock \emph{The International Journal of Robotics Research}, 02783649241273668.

\bibitem[{Driess et~al.(2023)Driess, Xia, Sajjadi, Lynch, Chowdhery, Wahid, Tompson, Vuong, Yu, Huang et~al.}]{palm_e}
Driess, D.; Xia, F.; Sajjadi, M.~S.; Lynch, C.; Chowdhery, A.; Wahid, A.; Tompson, J.; Vuong, Q.; Yu, T.; Huang, W.; et~al. 2023.
\newblock Palm-e: An embodied multimodal language model.

\bibitem[{Guo et~al.(2025)Guo, Zhang, Chen, Ji, Wang, Hu, and Chen}]{guo2025improving}
Guo, Y.; Zhang, J.; Chen, X.; Ji, X.; Wang, Y.-J.; Hu, Y.; and Chen, J. 2025.
\newblock Improving Vision-Language-Action Model with Online Reinforcement Learning.
\newblock \emph{arXiv preprint arXiv:2501.16664}.

\bibitem[{Ho, Jain, and Abbeel(2020)}]{DBLP:journals/corr/abs-2006-11239}
Ho, J.; Jain, A.; and Abbeel, P. 2020.
\newblock Denoising Diffusion Probabilistic Models.
\newblock \emph{CoRR}, abs/2006.11239.

\bibitem[{Hou et~al.(2025)Hou, Zhang, Xiong, Duan, Pu, Tong, Zhao, Zhu, Qiao, Dai et~al.}]{hou2025dita}
Hou, Z.; Zhang, T.; Xiong, Y.; Duan, H.; Pu, H.; Tong, R.; Zhao, C.; Zhu, X.; Qiao, Y.; Dai, J.; et~al. 2025.
\newblock Dita: Scaling diffusion transformer for generalist vision-language-action policy.
\newblock \emph{arXiv preprint arXiv:2503.19757}.

\bibitem[{Janner et~al.(2022)Janner, Du, Tenenbaum, and Levine}]{janner2022planning}
Janner, M.; Du, Y.; Tenenbaum, J.; and Levine, S. 2022.
\newblock Planning with Diffusion for Flexible Behavior Synthesis.
\newblock In \emph{International Conference on Machine Learning}, 9902--9915. PMLR.

\bibitem[{Kim et~al.(2024)Kim, Pertsch, Karamcheti, Xiao, Balakrishna, Nair, Rafailov, Foster, Lam, Sanketi et~al.}]{openvla}
Kim, M.~J.; Pertsch, K.; Karamcheti, S.; Xiao, T.; Balakrishna, A.; Nair, S.; Rafailov, R.; Foster, E.; Lam, G.; Sanketi, P.; et~al. 2024.
\newblock Openvla: An open-source vision-language-action model.
\newblock \emph{arXiv preprint arXiv:2406.09246}.

\bibitem[{Kool, van Hoof, and Welling(2019)}]{kool2019buy}
Kool, W.; van Hoof, H.; and Welling, M. 2019.
\newblock Buy 4 reinforce samples, get a baseline for free!

\bibitem[{Li et~al.(2024)Li, Hsu, Gu, Pertsch, Mees, Walke, Fu, Lunawat, Sieh, Kirmani et~al.}]{li2024evaluating}
Li, X.; Hsu, K.; Gu, J.; Pertsch, K.; Mees, O.; Walke, H.~R.; Fu, C.; Lunawat, I.; Sieh, I.; Kirmani, S.; et~al. 2024.
\newblock Evaluating real-world robot manipulation policies in simulation.
\newblock \emph{arXiv preprint arXiv:2405.05941}.

\bibitem[{Lipman et~al.(2023)Lipman, Chen, Ben-Hamu, Nickel, and Le}]{FM}
Lipman, Y.; Chen, R. T.~Q.; Ben-Hamu, H.; Nickel, M.; and Le, M. 2023.
\newblock Flow Matching for Generative Modeling.
\newblock arXiv:2210.02747.

\bibitem[{Liu et~al.(2023)Liu, Zhu, Gao, Feng, Liu, Zhu, and Stone}]{liu2023libero}
Liu, B.; Zhu, Y.; Gao, C.; Feng, Y.; Liu, Q.; Zhu, Y.; and Stone, P. 2023.
\newblock LIBERO: Benchmarking Knowledge Transfer for Lifelong Robot Learning.
\newblock \emph{arXiv preprint arXiv:2306.03310}.

\bibitem[{Loshchilov and Hutter(2017)}]{loshchilov2017decoupled}
Loshchilov, I.; and Hutter, F. 2017.
\newblock Decoupled weight decay regularization.
\newblock \emph{arXiv preprint arXiv:1711.05101}.

\bibitem[{Lu et~al.(2023)Lu, Chen, Chen, Su, Li, and Zhu}]{lu2023contrastive}
Lu, C.; Chen, H.; Chen, J.; Su, H.; Li, C.; and Zhu, J. 2023.
\newblock Contrastive energy prediction for exact energy-guided diffusion sampling in offline reinforcement learning.
\newblock In \emph{International Conference on Machine Learning}, 22825--22855. PMLR.

\bibitem[{Lu et~al.(2025)Lu, Guo, Zhang, Zhou, Jiang, Gao, Tang, and Wang}]{lu2025vla}
Lu, G.; Guo, W.; Zhang, C.; Zhou, Y.; Jiang, H.; Gao, Z.; Tang, Y.; and Wang, Z. 2025.
\newblock Vla-rl: Towards masterful and general robotic manipulation with scalable reinforcement learning.
\newblock \emph{arXiv preprint arXiv:2505.18719}.

\bibitem[{Mark et~al.(2024)Mark, Gao, Sampaio, Srirama, Sharma, Finn, and Kumar}]{mark2024policy}
Mark, M.~S.; Gao, T.; Sampaio, G.~G.; Srirama, M.~K.; Sharma, A.; Finn, C.; and Kumar, A. 2024.
\newblock Policy Agnostic RL: Offline RL and Online RL Fine-Tuning of Any Class and Backbone.
\newblock \emph{arXiv preprint arXiv:2412.06685}.

\bibitem[{Mete et~al.(2024)Mete, Xue, Wilcox, Chen, and Garg}]{mete2024quest}
Mete, A.; Xue, H.; Wilcox, A.; Chen, Y.; and Garg, A. 2024.
\newblock Quest: Self-supervised skill abstractions for learning continuous control.
\newblock \emph{Advances in Neural Information Processing Systems}, 37: 4062--4089.

\bibitem[{Nakamoto et~al.(2024)Nakamoto, Mees, Kumar, and Levine}]{nakamoto2024steering}
Nakamoto, M.; Mees, O.; Kumar, A.; and Levine, S. 2024.
\newblock Steering your generalists: Improving robotic foundation models via value guidance.
\newblock \emph{arXiv preprint arXiv:2410.13816}.

\bibitem[{Peters and Schaal(2007)}]{peters2007reinforcement}
Peters, J.; and Schaal, S. 2007.
\newblock Reinforcement learning by reward-weighted regression for operational space control.
\newblock In \emph{Proceedings of the 24th International Conference on Machine Learning}, ICML '07, 745–750. New York, NY, USA: Association for Computing Machinery.
\newblock ISBN 9781595937933.

\bibitem[{Reuss et~al.(2024)Reuss, Ya{\u{g}}murlu, Wenzel, and Lioutikov}]{reuss2024multimodal}
Reuss, M.; Ya{\u{g}}murlu, {\"O}.~E.; Wenzel, F.; and Lioutikov, R. 2024.
\newblock Multimodal diffusion transformer: Learning versatile behavior from multimodal goals.
\newblock \emph{arXiv preprint arXiv:2407.05996}.

\bibitem[{Schulman et~al.(2017)Schulman, Wolski, Dhariwal, Radford, and Klimov}]{schulman2017proximal}
Schulman, J.; Wolski, F.; Dhariwal, P.; Radford, A.; and Klimov, O. 2017.
\newblock Proximal policy optimization algorithms.
\newblock \emph{arXiv preprint arXiv:1707.06347}.

\bibitem[{Tan et~al.(2025)Tan, Dou, Zhao, and Kr{\"a}henb{\"u}hl}]{tan2025interactive}
Tan, S.; Dou, K.; Zhao, Y.; and Kr{\"a}henb{\"u}hl, P. 2025.
\newblock Interactive Post-Training for Vision-Language-Action Models.
\newblock \emph{arXiv preprint arXiv:2505.17016}.

\bibitem[{Team et~al.(2024)Team, Ghosh, Walke, Pertsch, Black, Mees, Dasari, Hejna, Kreiman, Xu et~al.}]{octo}
Team, O.~M.; Ghosh, D.; Walke, H.; Pertsch, K.; Black, K.; Mees, O.; Dasari, S.; Hejna, J.; Kreiman, T.; Xu, C.; et~al. 2024.
\newblock Octo: An open-source generalist robot policy.
\newblock \emph{arXiv preprint arXiv:2405.12213}.

\bibitem[{Wang, Hunt, and Zhou(2022)}]{wangdiffusion}
Wang, Z.; Hunt, J.~J.; and Zhou, M. 2022.
\newblock Diffusion Policies as an Expressive Policy Class for Offline Reinforcement Learning.
\newblock In \emph{The Eleventh International Conference on Learning Representations}.

\bibitem[{Zhai et~al.(2024)Zhai, Bai, Lin, Pan, Tong, Zhou, Suhr, Xie, LeCun, Ma et~al.}]{zhai2024fine}
Zhai, S.; Bai, H.; Lin, Z.; Pan, J.; Tong, P.; Zhou, Y.; Suhr, A.; Xie, S.; LeCun, Y.; Ma, Y.; et~al. 2024.
\newblock Fine-tuning large vision-language models as decision-making agents via reinforcement learning.
\newblock \emph{Advances in neural information processing systems}, 37: 110935--110971.

\bibitem[{Zhang et~al.(2025)Zhang, Zhuang, Zhao, Ding, Lu, and Wang}]{reinbot}
Zhang, H.; Zhuang, Z.; Zhao, H.; Ding, P.; Lu, H.; and Wang, D. 2025.
\newblock ReinboT: Amplifying Robot Visual-Language Manipulation with Reinforcement Learning.
\newblock \emph{arXiv preprint arXiv:2505.07395}.

\bibitem[{Zhang, Zhang, and Gu(2025)}]{EWFM}
Zhang, S.; Zhang, W.; and Gu, Q. 2025.
\newblock Energy-Weighted Flow Matching for Offline Reinforcement Learning.
\newblock arXiv:2503.04975.

\bibitem[{Zhang et~al.(2024)Zhang, Zheng, Chen, Jang, Li, Wang, Ding, Fox, and Yao}]{zhang2024grape}
Zhang, Z.; Zheng, K.; Chen, Z.; Jang, J.; Li, Y.; Wang, C.; Ding, M.; Fox, D.; and Yao, H. 2024.
\newblock Grape: Generalizing robot policy via preference alignment.
\newblock \emph{arXiv preprint arXiv:2411.19309}.

\bibitem[{Zhao et~al.(2025)Zhao, Song, Wang, Tong, Ding, Cheng, and Ge}]{zhao2025more}
Zhao, H.; Song, W.; Wang, D.; Tong, X.; Ding, P.; Cheng, X.; and Ge, Z. 2025.
\newblock MoRE: Unlocking Scalability in Reinforcement Learning for Quadruped Vision-Language-Action Models.
\newblock \emph{arXiv preprint arXiv:2503.08007}.

\bibitem[{Zheng et~al.(2023)Zheng, Le, Shaul, Lipman, Grover, and Chen}]{zheng2023guided}
Zheng, Q.; Le, M.; Shaul, N.; Lipman, Y.; Grover, A.; and Chen, R.~T. 2023.
\newblock Guided flows for generative modeling and decision making.
\newblock \emph{arXiv preprint arXiv:2311.13443}.

\bibitem[{Zhuang et~al.(2024)Zhuang, Peng, Liu, Zhang, and Wang}]{zhuang2024reinformer}
Zhuang, Z.; Peng, D.; Liu, J.; Zhang, Z.; and Wang, D. 2024.
\newblock Reinformer: Max-return sequence modeling for offline rl.
\newblock \emph{arXiv preprint arXiv:2405.08740}.

\end{thebibliography}

\clearpage
\section{Technical Appendix}
\subsection{Energy-weighted flow matching}
\label{app:Energy-Weighted Flow Matching}
The proof process of the energy-weighted flow matching theorem is given in this section. \\
\textbf{Theorem \ref{thm:1}} 
\textit{
Suppose $\qzero\propto\pzero\exp(-\beta\cE(\xb_0))$, and $\ptzero=\qtzero=\mathcal{N}(\alpha_t\xb_0,\sigma_t^2\mathbf{I})$. Consider the conditional vector field $\utzero$ which generates $\ptzero$, and the vector field $\uthat$ which generates $\qt$, then we have:\\
    \begin{align}
        \uthat = \int_{\xb_0} p_{0t}(\xb_0 | \xb) \utzero \frac{\exp(-\beta \cE(\xb_0))}{\exp(-\cE_t(\xb))}\mathrm d\xb_0,
    \end{align}
    where $\cE_t(\xb)$ is an intermediate energy function which is defined as: 
    $$\cE_t(\xb) = -\log \mathbb{E}_{p_{0t}(\xb_0|\xb)}[\exp(-\beta \cE(\xb_0))].$$
}
\textbf{Proof}: 
The integral form of the total probability formula of $\qt$ is:
$$\qt = \int_{\xb_0}\qtzero \mathrm d \xb_0.$$
Since $\pzero$ and $\qzero$ are probability distributions, $\int_{\xb_0} \qzero\mathrm d\xb_0=1$, so 
$$\qzero = \frac{\pzero\exp(-\beta\cE(\xb_0))}{\mathbb{E}_{\xb_0\sim \pzero}[\exp(-\beta\cE(\xb_0))]}.$$
So there is: 
\begin{align*}
\qt = &\int_{\xb_0}\qtzero \qzero\mathrm d \xb_0\\
=&\int_{\xb_0}\ptzero \frac{\pzero\exp(-\beta\cE(\xb_0))}{\mathbb{E}_{\xb_0\sim \pzero}[\exp(-\beta\cE(\xb_0))]} \mathrm d \xb_0\\
=&\int_{\xb_0}\pt p_{0t}(\xb_0|\xb) \frac{\exp(-\beta\cE(\xb_0))}{\mathbb{E}_{\xb_0\sim \pzero}[\exp(-\beta\cE(\xb_0))]} \mathrm d \xb_0 \quad (*)\\
=&\pt \frac{\mathbb{E}_{\xb_0\sim p_{0t}(\xb_0|\xb)}[\exp(-\beta \cE(\xb_0))]}{\mathbb{E}_{\xb_0\sim \pzero}[\exp(-\beta\cE(\xb_0))]}\\
=&\frac{\pt \exp(-\cE_t(\xb))}{\mathbb{E}_{\xb_0\sim \pzero}[\exp(-\beta\cE(\xb_0))]}.
\end{align*}
In (*) there is the Bayesian rule to have $\pt p_{0t}(\xb_0|\xb)=\ptzero\pzero$.\\
From \textbf{Definition \ref{def:1}}, there is: 
$$\frac{\mathrm{d}\ptzero}{\mathrm d t}= -\mathrm{div}(\ptzero\utzero).$$
While $\ptzero=\qtzero$, so 
\begin{align*}
    &\frac{\mathrm{d}\qt}{\mathrm d t}=\frac{\mathrm d}{\mathrm dt}\left(\frac{\pt\exp(-\cE_t(\xb))}{\mathbb{E}_{\xb_0\sim\pzero}[\exp(-\beta\cE(\xb_0))]}\right)\\
    =& \frac{\mathrm d}{\mathrm dt}\left(\frac{\pt \int_{\xb_0}p_{0t}(\xb_0|\xb)\exp(-\beta\cE(\xb_0))\mathrm d \xb_0}{\mathbb{E}_{\xb_0\sim\pzero}[\exp(-\beta\cE(\xb_0))]}\right)\\
    =&\frac{\mathrm d}{ \mathrm dt}\left(\frac{\int_{\xb_0}\pzero\ptzero\exp(-\beta\cE(\xb_0))\mathrm d \xb_0}{\mathbb{E}_{\xb_0\sim\pzero}[\exp(-\beta\cE(\xb_0))]}\right)\\
    =&\frac{\int_{\xb_0}\mathrm d \ptzero}{\mathrm d t}\frac{\pzero\exp(-\beta\cE(\xb_0))\mathrm d \xb_0}{\mathbb{E}_{\xb_0\sim\pzero}[\exp(-\beta\cE(\xb_0))]}\\
    =&-\int_{\xb_0}\mathrm{div}(\utzero\ptzero)\frac{\pzero\exp(-\beta\cE(\xb_0))\mathrm d \xb_0}{\mathbb{E}_{\xb_0\sim\pzero}[\exp(-\beta\cE(\xb_0))]}\\
    =&-\int_{\xb_0}\frac{\mathrm{div}\left(\utzero\ptzero\pzero\exp(-\beta\cE(\xb_0))\mathrm d \xb_0\right)}{\mathbb{E}_{\xb_0\sim\pzero}[\exp(-\beta\cE(\xb_0))]}\\
    =&-\int_{\xb_0}\mathrm{div}\left(\frac{\pt\utzero}{\mathbb{E}_{\xb_0\sim\pzero}[\exp(-\beta\cE(\xb_0))]}\right.\\
    &\left.\frac{ p_{0t}(\xb_0|\xb)\exp(-\beta\cE(\xb_0))\mathrm d \xb_0}{1}\right)\\
    =&-\int_{\xb_0}\mathrm{div}\left(\frac{\qt}{\exp(-\cE_t(\xb))}\right.\\
    &\left.\utzero p_{0t}(\xb_0|\xb)\exp(-\beta\cE(\xb_0))\mathrm d \xb_0\right)\\
    =&-\mathrm{div}\left(\qt\int\utzero p_{0t}(\xb_0|\xb)\frac{\exp(-\beta\cE(\xb_0))}{\exp(-\cE_t(\xb))}\mathrm d \xb_0\right).
\end{align*}
So $\hat{u}_t(\xb)=\int\utzero p_{0t}(\xb_0|\xb) \frac{\exp(-\beta\cE(\xb_0))}{\exp(-\cE_t(\xb))}\mathrm d \xb_0$. \\
\textbf{Theorem \ref{thm:2}} 
\textit{
$p_0(\xb_0),q_0(\xb_0),\pt,\qt$$,\cE(\xb_0),\cE_t(\xb_t), \\ \uthat,\utzero$ are all defined above. Consider learning a model $\vb_{\theta}(\xb)$ with the learnable parameter $\theta$, then define the Energy-weighted Flow Matching (EFM) loss $\mathcal{L}_{EFM}$ as:
$$\mathcal{L}_{EFM}(\theta)=\mathbb{E}_{\xb,t}[\frac{\exp(- \cE_t(\xb))}{\mathbb{E}_{\tilde{x}\sim p_t(\tilde \xb)}[\exp(- \cE_t(\tilde \xb))]}||\vb_{\theta}(\xb)-\uthat||^2],$$
and the Conditional Energy-weighted Flow Matching (CEFM) loss $\mathcal{L}_{CEFM}(\theta)=$:
$$\mathbb{E}_{\xb_0,\xb,t}[\frac{\exp(-\beta\cE(\xb_0))}{\mathbb{E}_{\tilde{x}\sim p_0(\tilde \xb_0)}[\exp(-\beta\cE(\tilde \xb_0))]}||\vb_{\theta}(\xb)-\utzero||^2].$$
Then there is $\nabla_{\theta}\mathcal{L}_{EFM}(\theta)=\nabla_{\theta}\mathcal{L}_{CEFM}(\theta)$.\\
}
\textbf{Proof}: The theorem is proved in three steps. Firstly, we prove the relationship between $\cE$ and $\cE_t$, that is:
$$\mathbb{E}_{\xb}[\exp(-\cE_t(\xb))]=\mathbb{E}_{\xb_0}[\exp(-\beta\cE(\xb_0))].$$
Secondly, we find the relationship between $||\vb_{\theta}(\xb)-\uthat||^2$ and $||\vb_{\theta}(\xb)-\utzero||^2$.
Lastly, we combine the previous findings and complete the whole proof.\\
For the first step, there is: 
\begin{align*}
&\mathbb{E}_{\xb}[\exp(-\cE_t(\xb)]\\
=&\int_{\xb}\pt\exp(-\cE_t(\xb))\mathrm d \xb\\
=&\int_{\xb}\pt\int_{\xb_0}p_{0t}(\xb_0|\xb)\exp(-\beta\cE_t(\xb_0))\mathrm d \xb_0\mathrm d \xb\\
=&\int_{\xb}\int_{\xb_0}p_{0t}(\xb_0|\xb)\pt\exp(-\beta\cE_t(\xb_0))\mathrm d \xb_0\mathrm d \xb\\
=&\int_{\xb}\int_{\xb_0}\pzero\ptzero\exp(-\beta\cE_t(\xb_0))\mathrm d \xb_0\mathrm d \xb\\
=&\int_{\xb_0}\left(\pzero\exp(-\beta\cE_t(\xb_0))\mathrm d \xb_0\int_{\xb}\ptzero\mathrm d \xb\right)\\
=&\int_{\xb_0}\pzero\exp(-\beta\cE_t(\xb_0))\mathrm d \xb_0.\\
=&\mathbb{E}_{\xb_0}[\exp(-\beta\cE(\xb_0))].
\end{align*}
For the second step, there is:
\begin{align*}
    &\nabla_{\theta}||\vb_{\theta}(\xb)-\hat{u}_t(\xb)||^2\\
    =&\nabla_{\theta}||\vb_{\theta}(\xb)||^2-2\nabla_{\theta}\langle\vb_{\theta}(\xb),\hat{u}_t(\xb)\rangle+\nabla_{\theta}||\hat{u}_t(\xb)||^2\\
    =&\nabla_{\theta}||\vb_{\theta}(\xb)||^2-2\nabla_{\theta}\langle\vb_{\theta}(\xb),\hat{u}_t(\xb)\rangle.\\
\end{align*}

So we have:\\
\begin{align*}
&\int_\xb\pt\frac{\exp(-\cE_t(\xb))}{\mathbb{E}[\exp(-\cE_t(\tilde{\xb}))]}\nabla_{\theta}||\vb_{\theta}(\xb)||^2\mathrm d\xb\\
=&\int_\xb\pt\frac{\exp(-\cE_t(\xb))}{\mathbb{E}[\exp(-\beta\cE(\tilde{\xb_0}))]}\nabla_{\theta}||\vb_{\theta}(\xb)||^2\mathrm d\xb\\
=&\int_\xb \qt\nabla_{\theta}||\vb_{\theta}(\xb)||^2\mathrm d\xb\\
=&\int_\xb\int_{\xb_0}\qtzero\qzero\mathrm d \xb_0\nabla_{\theta}||\vb_{\theta}(\xb)||^2\mathrm d\xb\\
=&\int_\xb\int_{\xb_0}\ptzero\pzero\frac{\exp(-\beta\cE(\xb_0))}{\mathbb{E}[\exp(-\beta\cE(\tilde{\xb}_0)]}\nabla_{\theta}||\vb_{\theta}(\xb)||^2\mathrm d\xb \mathrm d \xb_0\\
=&\mathbb{E}_{\xb_0,\xb,t}[\frac{\exp(-\beta\cE(\xb_0))}{\mathbb{E}[\exp(-\beta\cE(\tilde{\xb}_0)]}||\vb_{\theta}(\xb)||^2].\\
\end{align*}
\begin{align*}
&\int_\xb\pt\frac{\exp(-\cE_t(\xb))}{\mathbb{E}[\exp(-\cE_t(\tilde{\xb}))]}\nabla_{\theta}\langle\vb_{\theta}(\xb),\hat{u}_t(\xb)\rangle\mathrm d\xb\\
=&\int_\xb\pt\frac{\exp(-\cE_t(\xb))}{\mathbb{E}[\exp(-\beta\cE(\tilde{\xb}_0))]}\mathrm d\xb\\
&\nabla_{\theta}\left\langle \vb_{\theta}(\xb),\int_{\xb_0}\utzero p_{0t}(\xb_0|\xb)\frac{\exp(-\beta\cE(\xb_0))}{\exp(-\cE_t(\xb))}\mathrm d\xb_0\right\rangle\\
=&\int_\xb\int_{\xb_0}\ptzero\pzero\frac{\exp(-\beta\cE(\xb_0))}{\mathbb{E}[\exp(-\beta\cE(\tilde{\xb}_0)]}\\
&\nabla_{\theta}\langle\vb_{\theta}(\xb),\utzero\rangle\mathrm d\xb\mathrm d\xb_0\\
=&\mathbb{E}_{\xb_0,\xb,t}[\frac{\exp(-\beta\cE(\xb_0))}{\mathbb{E}[\exp(-\beta\cE(\tilde{\xb}_0)]}\nabla_{\theta}\langle\vb_{\theta}(\xb),\utzero\rangle].
\end{align*}
In the last step, we combine the previous three equations, we have:
\begin{align*}
    &\nabla_{\theta}\left(\mathbb{E}_{\xb,t}[\frac{\exp(- \cE_t(\xb))}{\mathbb{E}_{\tilde{x}\sim p_t(\tilde \xb)}[\exp(- \cE_t(\tilde \xb))]}||\vb_{\theta}(\xb)-\hat{u}_t(\xb)||^2\right)\\
    =&\mathbb{E}_{\xb_0,\xb,t}[\frac{\exp(-\beta\cE(\xb_0))}{\mathbb{E}[\exp(-\beta\cE(\tilde{\xb}_0)]}\nabla_{\theta}||\vb_{\theta}(\xb)||^2]\\
    -2&\mathbb{E}_{\xb_0,\xb,t}[\frac{\exp(-\beta\cE(\xb_0))}{\mathbb{E}[\exp(-\beta\cE(\tilde{\xb}_0)]}\nabla_{\theta}\langle\vb_{\theta}(\xb),\utzero\rangle].\\
    =&\nabla_{\theta}\left(\mathbb{E}_{\xb_0,\xb,t}[\frac{\exp(-\beta\cE(\xb_0))}{\mathbb{E}[\exp(-\beta\cE(\tilde{\xb}_0)]}||\vb_{\theta}(\xb)-\utzero||^2\right).\\
\end{align*}
So that is $\nabla_{\theta}\mathcal{L}_{EFM}(\theta)=\nabla_{\theta}\mathcal{L}_{CEFM}(\theta)$.

\subsection{Derivation of tuning process of $\alpha$} 
\label{app:Derivation of tuning process}
In this section, we discuss the derivation of \textbf{Corollary}~\ref{cor:alpha1} and \textbf{Corollary}~\ref{cor:alpha2}, which are about the dynamic tuning process of the scaling factor $\alpha$.\\
\textbf{Corollary \ref{cor:alpha1}}
\textit{
Define $J(\alpha) = \mathrm{Var}(\hat{g}(\alpha))-\lambda S(\alpha),$ then $J(\alpha)= \sigma_L^2[e^{2\alpha^2\sigma_R^2}-e^{\alpha^2\sigma_R^2}]-\lambda \alpha \sigma_R^2$.\\
}
\textbf{Proof}: By the assumption, all of the $R^{*}(\Ab_t^j,\ob_t)$ satisfy the same distribution $R^* \sim \mathcal{N}(0,\sigma_R^2)$, so $\hat{w}_i = \exp(\alpha R^*(\Ab_t^i,\ob_t^i))$ are also i.i.d. variables. 
Then we define $m_1(\alpha)=\mathbb{E}[\exp(\alpha R^*)]$, and $m_2(\alpha) = \mathbb{E}[\exp(2\alpha R^*)]$.

Firstly, we derive $m_1(\alpha)$ and $m_2(\alpha)$:
\begin{align*}
&m_1(\alpha)=\mathbb{E}[\exp(\alpha R^*)]\\
&=\int_{-\infty}^{+\infty}\exp(\alpha x)\mathrm{Pr}(R^*=x)\mathrm d x\\
&=\int_{-\infty}^{+\infty}\frac{1}{\sqrt{2\pi\sigma_R}}\exp(\alpha x)\exp(-\frac{x^2}{2\sigma_R^2})\mathrm d x\\
&=\int_{-\infty}^{+\infty}\frac{1}{\sqrt{2\pi\sigma_R}}\exp(-\frac{1}{2\sigma_R^2}(x-\sigma_R\alpha)^2)\exp(\frac{1}{2}\sigma_R^2\alpha^2)\mathrm d x\\
&=\exp(\frac{1}{2}\sigma_R^2\alpha^2),\\
&m_2(\alpha)=\mathbb{E}[\exp(2\alpha R^*)]\\
&=m_1(2\alpha)=\exp(\frac{1}{2}4\sigma_R^2\alpha^2)=\exp(2\alpha^2\sigma_R^2).\\
\end{align*}
Secondly, we start to derive $\mathrm{Var}(\hat{g}(\alpha))$ and $S(\alpha)$: \\
\begin{align*}
\mathrm{Var}(\hat{g}(\alpha))=&\mathrm{Var}\left(\frac{1}{B}\sum_{i=1}^B\hat{w}_i(\alpha)\nabla_{\theta}\mathcal{L}_{CFM}^i\right)\\
=&\frac{1}{B}\sum_{i=1}^B\left(\mathbb{E}[\exp(\alpha R_i^*)(\nabla_{\theta}\mathcal{L}_{CFM}^i)^2]\right.\\
&\left.-(\mathbb{E}[\exp(\alpha R_i^*)\nabla_{\theta}\mathcal{L}_{CFM}^i])^2\right)\\
=&\frac{1}{B}\sum_{i=1}^B(m_2(\alpha)-(m_1(\alpha))^2)\sigma_L^2\\
=&(\exp(2\alpha^2\sigma_R^2)-\exp(\alpha^2\sigma_R^2))\sigma_L^2,\\
S(\alpha)=&(B\mathbb{E}[\exp(\alpha R_i^*)R_i^*]) / (B \mathbb{E}[\exp(\alpha R_i^*)])\\
=&\int_{-\infty}^{+\infty}\mathrm{Pr}(R^*_i=x)\frac{1}{m_1(\alpha)}\exp(\alpha x)x\mathrm dx\\
=&\int_{-\infty}^{+\infty}\frac{1}{m(\alpha)}\mathrm{Pr}(R^*_i=x)\frac{\mathrm d}{\mathrm d\alpha}(\exp(\alpha x))\mathrm dx\\
=&\frac{\mathrm d}{m_1(\alpha)\mathrm d\alpha}\left(\int_{-\infty}^{+\infty}\mathrm{Pr}(R^*_i=x)(\exp(\alpha x))\mathrm dx\right)\\
=&\frac{m_1'(\alpha)}{m_1(\alpha)}\\
=&\frac{\alpha\sigma_R^2\exp(\frac{1}{2}\alpha^2\sigma_R^2)}{\exp(\frac{1}{2}\alpha^2\sigma_R^2)}\\
=&\alpha\sigma_R^2.
\end{align*}
\textbf{Corollary \ref{cor:alpha2}} 
\textit{
The optimal $\alpha^*$ which minimizes $J(\alpha)$ is solved by:\\
\begin{align}
       4\sqrt{x^*}e^{2x^*}-2\sqrt{x^*}e^{x^*}-\frac{\lambda\sigma_R}{\sigma_L^2} = 0, \alpha^*=\frac{\sqrt{x^*}}{\sigma_R}.
\end{align}
}
\textbf{Proof}: We need to solve $J'(\alpha)=0$.
\begin{align*}
    &J'(\alpha)=0\\
    &\Rightarrow \frac{\mathrm d J}{\mathrm d \alpha}=\sigma_L^2[4\alpha^2\exp(2\alpha^2\sigma_R^2)-2\alpha \sigma_R^2\exp(\alpha^2\sigma_R^2)] - \lambda \sigma_R^2=0\\
    & \Rightarrow 4\alpha^2\exp(2\alpha^2\sigma_R^2)-2\alpha\exp(\alpha^2\sigma_R^2)-\frac{\lambda}{\sigma_L^2} = 0.\\
    &\mathrm{Let}\ x=\alpha^2\sigma_R^2, 4\sqrt{x}\exp(2x) - 2\sqrt{x}\exp(x) - \frac{\lambda\sigma_R}{\sigma_L^2}=0.
\end{align*}
So $\alpha^* = \frac{\sqrt{x^*}}{\sigma_R}$.\\

\section{Data Appendix}
To thoroughly evaluate the performance of the proposed ARFM approach, we conduct comparative experiments in multiple experimental settings on the LIBERO~\cite{liu2023libero} simulation and the real UR5 platform. LIBERO is a comprehensive lifelong learning benchmark that contains multiple task suites. These suites are designed to evaluate specific aspects of general robotic manipulation and define tasks through language-guided instructions. Specifically, LIBERO contains four main suites: object, long, space, and goal. Each suite is customized to test different object manipulation capabilities and contains 10 tasks. 

To evaluate the performance of ARFM in the real world, we collected approximately 720 successful trajectories (approximately 34,600 frames) on a UR5 robot arm in the real world, with each trajectory length varying from several tens of seconds. The tasks primarily involved grasping and placing objects such as white squares, corn, cups, and peppers. The data includes first- and third-person RGB images ($480\times640\times3$ dimensions), robot joint angles (7 dimensions), and robot desired joint angles (7 dimensions).

In terms of reward function design in the dataset, previous work $--$ ReinboT~\cite{reinbot} considered the need for VLA model policies to maintain robust and stable behavior with minimal energy costs while following goals. Therefore, ReinboT considers four aspects of reward components to dense rewards: sub-goal achievement, task progress, behavior smoothness and task completion. 
Following this dense reward design principle, the dense rewards we utilized in this work include 13 reward components. 
The specific reward items and weights are shown in Tab.~\ref{tab:rewards}.
The dataset we use will be publicly available upon publication.

\section{Code Appendix}
\label{app:Implementation Details}
The ARFM we propose and the reproduced flow matching type baseline algorithm ($\pi_0$, ReinboT, RWR) are all implemented based on the code repository: \textit{huggingface/lerobot}~\cite{cadene2024lerobot}.
The hyperparameter configuration is shown in Tab.~\ref{tab:trainer}. 
The ARFM method (and the reproduced baseline) utilizes 2 GPUs, and 40,000 steps of fine-tuning the full parameters of the four types of task data of LIBERO, which takes about 11 hours; 60,000 steps of fine-tuning the realistic UR5 manipulation data (more than 700 success trajectories), which takes about 16 hours. 
CPU: Intel(R) Xeon(R) Platinum 8358 CPU @ 2.60GHz. 
GPU: NVIDIA A100-SXM4-80GB.

For the reproduction of the ReinboT baseline algorithm of flow matching type, we splice the learnable tokens on the Gemma Expert module of $\pi_0$, and utilize the expected regression loss to learn return-to-go. 
Then, the gate mechanism~\cite{alayrac2022flamingo} is utilized to fuse the predicted return-to-go encoding and the action vector field encoding of $\pi_0$, and then predict the action vector field through flow matching loss. 
The code we use will be publicly available upon publication.

\begin{figure*}[!th]
  \centering
  \begin{subfigure}[t]{0.5\textwidth}
    \centering
    \includegraphics[width=0.9\linewidth]{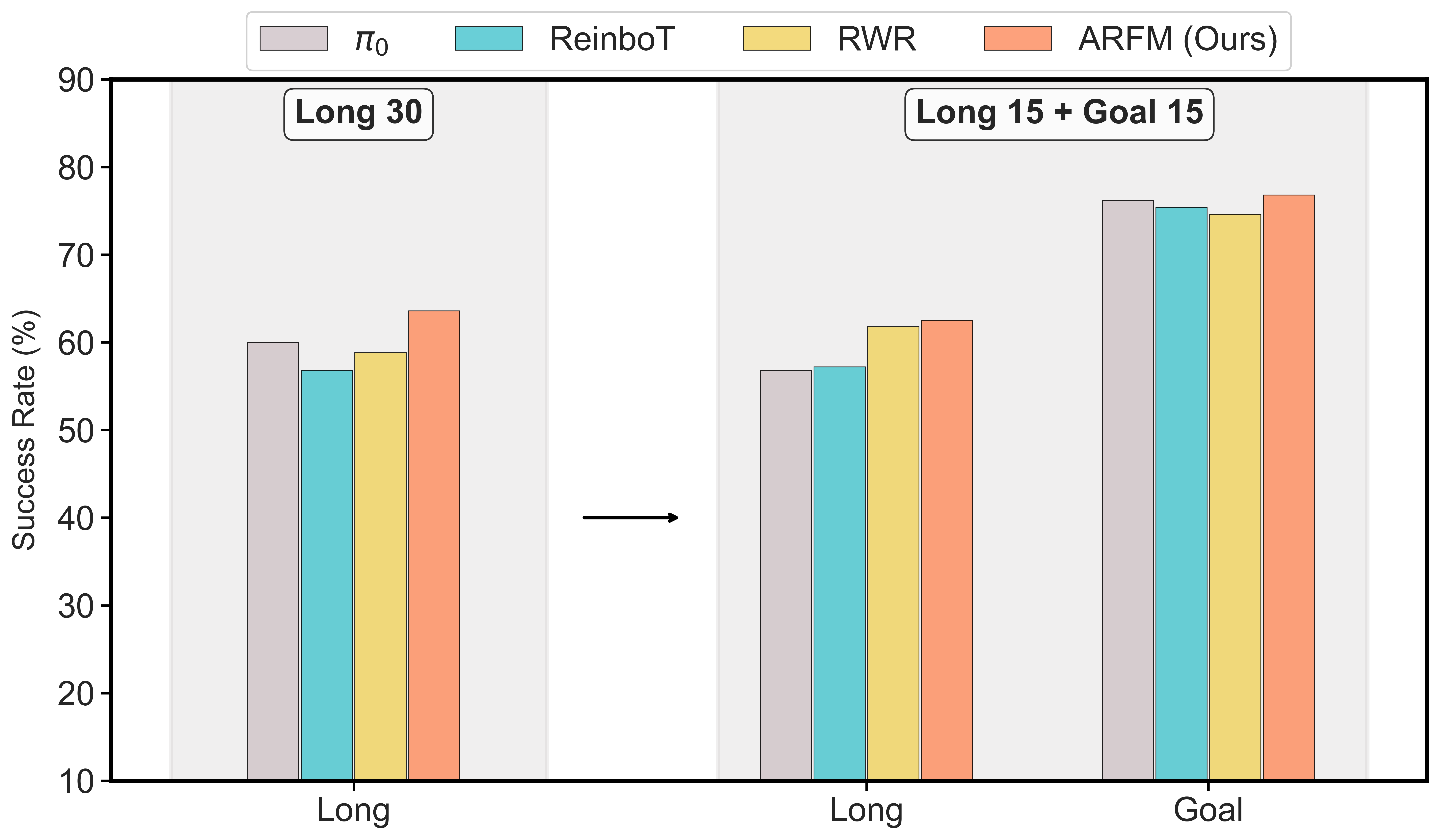}
    \caption{}
    \label{fig:l30_l15_g15}
  \end{subfigure}
  \hfill
  \begin{subfigure}[t]{0.45\textwidth}
    \centering
    \includegraphics[width=0.9\linewidth]{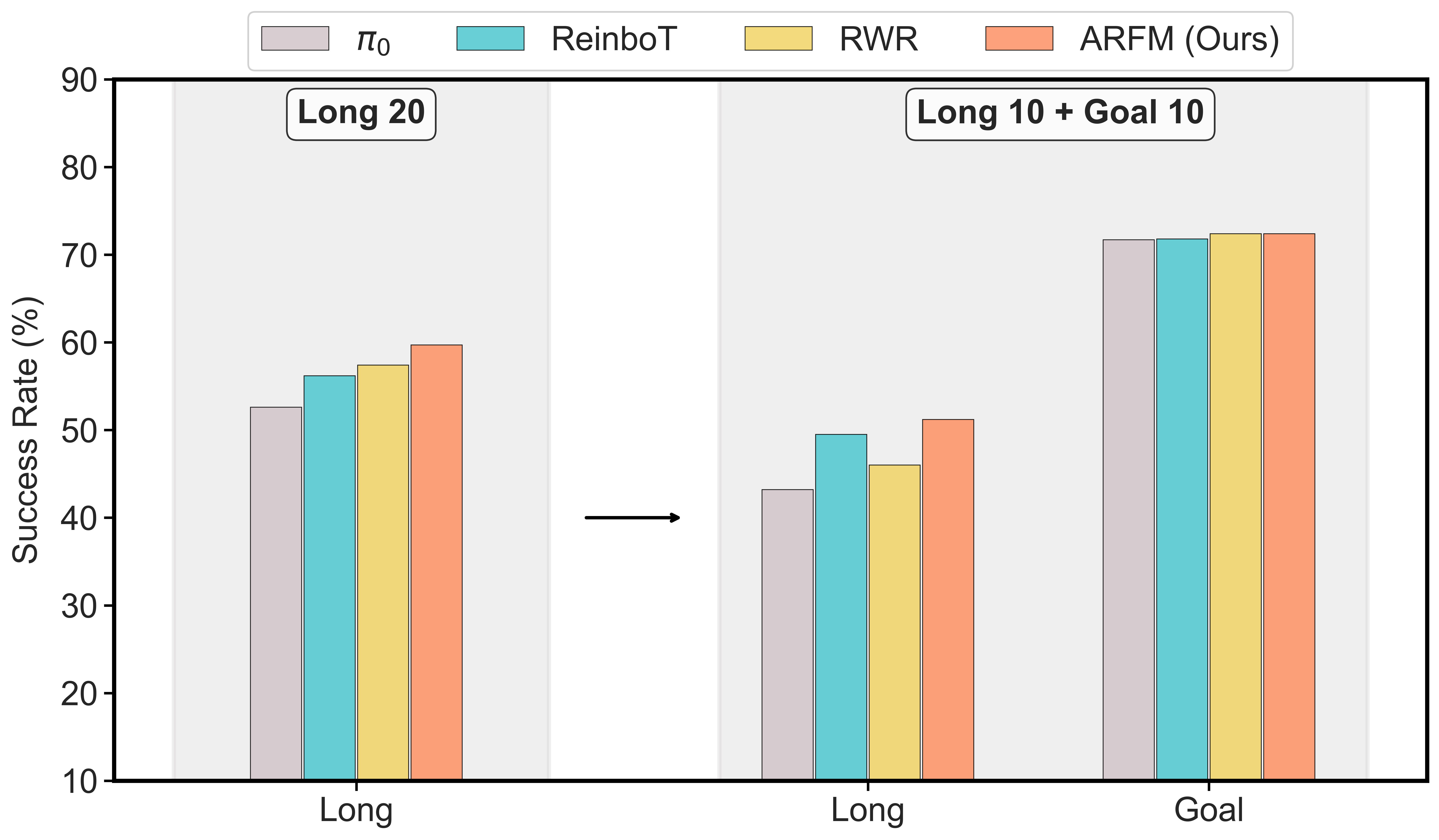}
    \caption{}
    \label{fig:l20_l10_g10}
  \end{subfigure}
  \caption{Comparison of continuous learning performance under two task compositions (noise levels=0.1). (a) Training sequence: Long(30) → Long(15) + Goal(15). (b) Training sequence: Long(20) → Long(10) + Goal(10).}
  \label{fig:continual_comparison_2}
\end{figure*}

\begin{table*}[!ht]

    \begin{center}
        \renewcommand\arraystretch{1.2}
        \resizebox{\linewidth}{!}{
        \begin{tabular}{c|cccc|cccc|cccc}
        \hline \thickhline
        \multicolumn{1}{c|}{\textbf{Few-Shot}}  
        & \multicolumn{4}{c|}{\textbf{30-shot}} 
        & \multicolumn{4}{c|}{\textbf{20-shot}} 
        & \multicolumn{4}{c}{\textbf{10-shot}} \\
        \hline
        \textbf{Action Noise} & 
        \textbf{0.10} & \textbf{0.15} & \textbf{0.20} & \textbf{Average} &  
       \textbf{0.10} & \textbf{0.15} & \textbf{0.20} & \textbf{Average} &  
        \textbf{0.10} & \textbf{0.15} & \textbf{0.20} & \textbf{Average} \\
        \hline \hline
        $\pi_0$ & 
        60.0 & 40.4 & 24.8 & 41.7 & 
        52.6 & 34.4 & 14.3 & 33.8 & 
        36.6 & 20.4 & 9.3 & 22.1 \\
        
        ReinboT & 
        56.8 & 38.8 & 22.8 & 39.5 &
        56.2 & 36.2 & 20.2 & 37.5 & 
        40.8 & 22.0 & 11.0 & 24.6 \\
        
        RWR & 
        58.8 & 38.8 & 21.0 & 39.5 & 
        57.4 & 34.6 & \textbf{21.1} & 37.7 & 
        41.2 & 27.2 & 11.6 & 26.7 \\
        
        \textbf{ARFM (Ours)} & 
        \textbf{63.6} & \textbf{41.6} & \textbf{23.6} & \textbf{42.9} & 
        \textbf{59.7} & \textbf{37.0} & 20.0 & \textbf{38.9} & 
        \textbf{42.4} & \textbf{27.3} & \textbf{13.4} & \textbf{27.7}  \\
        \hline \thickhline
        \end{tabular}
        }
    \end{center}
    \caption{
    Comparison of model SR(\%) under few-shot learning settings on the LIBERO-Long suite.
    }
    \label{table:libero-few-shot-all}
\end{table*}

\begin{table*}[!ht]

    \begin{center}
        \renewcommand\arraystretch{1.2}

        \resizebox{\linewidth}{!}{
        \begin{tabular}{c|c|cc|cc|c|cc|cc}
        \hline \thickhline
        \multirow{3}{*}{\textbf{Models}} & 
        \multicolumn{10}{c}{\textbf{LIBERO Continual Learning}}  \\
        \cline{2-11}
        & \multicolumn{5}{c|}{\textbf{L 30 $\rightarrow$ L 15 + G 15}} & 
        \multicolumn{5}{c}{\textbf{L 20 $\rightarrow$ L 10 + G 10}} \\
        \cline{2-11}
        & \textbf{Long} & \textbf{Long} & \textbf{Goal} & \textbf{Long NBT \(\downarrow\)} & \textbf{Avg. SR \(\uparrow\)} 
        & \textbf{Long} & \textbf{Long} & \textbf{Goal} & \textbf{Long NBT \(\downarrow\)} & \textbf{Avg. SR \(\uparrow\)} \\
        \hline \hline
        $\pi_0$ & 
        60.0 & 56.8 & 76.2 & 3.2 & 64.3 & 
        52.6 & 43.2 & 71.7 & 9.4 & 55.8 \\
        
        ReinboT & 
        56.8 & 57.2 & 75.4 & \textbf{0} & 63.1 & 
        56.2 & 49.5 & 71.8 & \textbf{6.7} & 59.2 \\
        
        RWR & 
        58.8 & 61.8 & 74.6 & \textbf{0} & 60.3 & 
        57.4 & 46.0 & \textbf{72.4} & 11.4 & 58.6 \\
        
        \textbf{ARFM (Ours)} & 
        \textbf{63.6} & \textbf{62.5} & \textbf{76.8} & 1.1 & \textbf{67.6} & 
        \textbf{59.7} & \textbf{51.2} & \textbf{72.4} & 8.5 & \textbf{61.1} \\
        \hline 
        \end{tabular}
        }

        \vspace{0.1em}  

        \resizebox{\linewidth}{!}{
        \begin{tabular}{>{\centering\arraybackslash}p{1.84cm}|c|cc|ccc|ccc}
        \hline 
        \multirow{2}{*}{\textbf{Models}} & 
        \multicolumn{9}{c}{\textbf{L 10 $\rightarrow$ L 5 + G 5 $\rightarrow$ L 2 + G 2 + O 2}} \\
        \cline{2-10}
        & \textbf{Long} & \textbf{Long} & \textbf{Goal} & \textbf{Long} & \textbf{Goal} & \textbf{Object} & \textbf{Long NBT \(\downarrow\)} & \textbf{Goal NBT \(\downarrow\)} & \textbf{Avg. SR \(\uparrow\)} \\
        \hline \hline
        $\pi_0$ & 
        36.6 & 33.2 & 68.6 & 26.0 & 61.8 & 
        47.2 & 10.6 & 6.8 & 45.6  \\
        
        ReinboT & 
        40.8 & 36.0 & 65.2 & 27.1 & 59.2 & 
        44.0 & 13.7 & 6.0 & 45.4  \\
        
        RWR & 
        41.2 & 40.3 & 66.2 & 26.2 & 62.8 & 
        46.0 & 14.5 & \textbf{3.4} & 47.1  \\
        
        \scalebox{0.9}{\textbf{ARFM (Ours)}} & 
        \textbf{42.4} & \textbf{53.1} & \textbf{74.0} & \textbf{39.0} & 
        \textbf{68.4} & \textbf{48.0} & \textbf{3.4} & 5.6 & \textbf{54.2}  \\
        \hline \thickhline
        \end{tabular}
        }

    \end{center}
    \caption{
    Model performance under continual learning with action noise = 0.10 on the LIBERO benchmark.
    }
    \label{table:libero-merged}
\end{table*}

\begin{table*}[htbp]
    \centering
    \begin{tabular*}{0.6\textwidth}{c @{\extracolsep{\fill}} c}
        \toprule
        \textbf{Parameter} & \textbf{Value} \\
        \midrule
        \multicolumn{2}{c}{\textit{General}} \\
        \midrule
        Total Post-Training Steps & $4e4$ (LIBERO); $6e4$ (UR5)\\
        Batch Size & 16 \\
        Action Horizon & 50 \\
        Optimization Objection Hyperparameter $\lambda$ & $5.0e-4$ \\
        Number of Bisection Iterations $M$ & 20 \\
        Value Range of $\alpha$ $[\alpha_{min},\alpha_{max}]$ & $[0.01,5]$ \\
        Tolerance $\epsilon$ & $1.0e-5$ \\
        \midrule
        \multicolumn{2}{c}{\textit{Optimizer (AdamW~\cite{loshchilov2017decoupled})}} \\
        \midrule
        Learning Rate & $1.0e-4$ \\
        Betas ($\beta_1, \beta_2$) & (0.9, 0.95) \\
        Epsilon & $1.0e-8$ \\
        Weight Decay & $1.0e-10$ \\ 
        Gradient Clip Norm & $10$ \\
        \midrule
        \multicolumn{2}{c}{\textit{Scheduler (Cosine Decay with Warmup)}} \\
        \midrule
        Warmup Steps & $1e3$ \\
        Decay Steps & $3e4$ \\
        Peak Learning Rate & $2.5e-5$ \\
        Decay Learning Rate & $2.5e-6$ \\
        \midrule
        \multicolumn{2}{c}{\textit{Image Augmentation}} \\
        \midrule
        Brightness Range & $[0.8, 1.2]$ \\
        Contrast Range & $[0.8, 1.2]$ \\
        Saturation Range & $[0.5, 1.5]$ \\
        Hue Range & $[-0.05, 0.05]$ \\
        Sharpness Range & $[0.5, 1.5]$ \\
        \bottomrule
    \end{tabular*}
     \caption{Training hyperparameters configuration.}
     \label{tab:trainer}
\end{table*}

\begin{table*}[htbp]
    \centering
    \begin{tabular*}{0.8\textwidth}{c @{\extracolsep{\fill}} c}
        \toprule
        \textbf{Reward Component} & \textbf{Weight} \\
        \midrule
        \multicolumn{2}{c}{\textit{Sub-goal Achievement}} \\
        \midrule
        Image MSE ($e^{f_{\rm \text{MSE}}(o_{t},o_{t}^{\ast})}$)& $0.1/13$ \\
        Image SSIM ($e^{f_{\rm \text{SSIM}}(o_{t},o_{t}^{\ast})}$)& $0.1/13$ \\
        Image ORB Similarity ($e^{f_{\rm \text{ORB}}(o_{t},o_{t}^{\ast})}$)& $0.1/13$ \\
        Gripper Image MSE ($e^{f_{\rm \text{MSE}}(o_{t},o_{t}^{\ast})}$)& $0.1/13$ \\
        Gripper Image SSIM ($e^{f_{\rm \text{SSIM}}(o_{t},o_{t}^{\ast})}$)& $0.1/13$ \\
        Gripper Image ORB Similarity ($e^{f_{\rm \text{ORB}}(o_{t},o_{t}^{\ast})}$)& $0.1/13$ \\
        Joint Position MSE ($e^{f_{\rm \text{MSE}}(s_{t},s_{t}^{\ast})}$)& $0.1/13$ \\
        \midrule
        \multicolumn{2}{c}{\textit{Task Progress}} \\
        \midrule
        Sub-goal Division ($\frac{n(s_t)}{\vert \{ s^{*} \} \vert}$)& $0.1/13$ \\
        \midrule
        \multicolumn{2}{c}{\textit{Behavior Smoothness}} \\
        \midrule
        Joint Velocity ($-|\dot{\mathbf{q}}|^2$)& $0.1/13$ \\
        Joint Acceleration ($-|\ddot{\mathbf{q}}|^2$)& $0.1/13$ \\
        Action Velocity ($-|\mathbf{a}_{t-1}-\mathbf{a}_{t}|^2$)& $0.01/13$ \\
        Action Acceleration ($-|\mathbf{a}_{t-2}-2\mathbf{a}_{t-1}+\mathbf{a}_t|^2$)& $0.01/13$ \\
        \midrule
        \multicolumn{2}{c}{\textit{Task Completion}} \\
        \midrule
        Task Success ($\mathbb{I}\{\tau\ \text{is}\ \text{successful}\}$)& $0.1/13$ \\
        \bottomrule
    \end{tabular*}
      \caption{Dense reward components and weights.}
      \label{tab:rewards}
\end{table*}





\end{document}